\documentclass[letterpaper]{article} 
\usepackage{aaai24}  
\usepackage{times}  
\usepackage{helvet}  
\usepackage{courier}  
\usepackage[hyphens]{url}  
\usepackage{graphicx} 
\usepackage{natbib}  
\usepackage{caption} 
\usepackage{algorithm}
\usepackage{algorithmic}

\usepackage{newfloat}
\usepackage{listings}
\DeclareCaptionStyle{ruled}{labelfont=normalfont,labelsep=colon,strut=off} 
\floatstyle{ruled}
\newfloat{listing}{tb}{lst}{}
\floatname{listing}{Listing}
\newif\ifincludeappendix
\includeappendixtrue  

\title{GLOP: Learning Global Partition and Local Construction\\ for Solving Large-scale Routing Problems in Real-time}
\author{
    Haoran Ye\textsuperscript{\rm 1}, Jiarui Wang\textsuperscript{\rm 1}, Helan Liang\textsuperscript{\rm 1}\thanks{Corresponding author.}, Zhiguang Cao\textsuperscript{\rm 2}, Yong Li\textsuperscript{\rm 3}, Fanzhang Li\textsuperscript{\rm 1}\\
}
\affiliations{
    \textsuperscript{\rm 1}Soochow University, China\\
    \textsuperscript{\rm 2}Singapore Management University, Singapore\\
    \textsuperscript{\rm 3}Tsinghua University, China\\
      \texttt{\{hrye,jrwangfurffico\}@stu.suda.edu.cn, \{hlliang,lfzh\}@suda.edu.cn}\\\texttt{zgcao@smu.edu.sg, liyong07@tsinghua.edu.cn}

}

\usepackage{bibentry}

\usepackage{xcolor}
\usepackage{bm}
\usepackage{multirow}
\usepackage{booktabs}
\usepackage{amsmath}
\usepackage{amssymb}

\begin{document}

\maketitle

\begin{abstract}
The recent end-to-end neural solvers have shown promise for small-scale routing problems but suffered from limited real-time scaling-up performance. This paper proposes GLOP (\textbf{G}lobal and \textbf{L}ocal \textbf{O}ptimization \textbf{P}olicies), a unified hierarchical framework that efficiently scales toward large-scale routing problems. GLOP partitions large routing problems into Travelling Salesman Problems (TSPs) and TSPs into Shortest Hamiltonian Path Problems. For the first time, we hybridize non-autoregressive neural heuristics for coarse-grained problem partitions and autoregressive neural heuristics for fine-grained route constructions, leveraging the scalability of the former and the meticulousness of the latter. Experimental results show that GLOP achieves competitive and state-of-the-art real-time performance on large-scale routing problems, including TSP, ATSP, CVRP, and PCTSP. Our code is available: \url{https://github.com/henry-yeh/GLOP}.

\end{abstract}

\section{Introduction}
Routing problems pervade logistics, supply chain, transportation, robotic systems, etc. Modern industries have witnessed ever-increasing demands for the massive and expeditious routing of goods, services, and people. Traditional solvers based on mathematical programming or iterative heuristics struggle to keep pace with such growing complexity and real-time requirements.

Recent advances in Neural Combinatorial Optimization (NCO) \cite{survey1} seek end-to-end solutions for routing problems, where neural solvers are exploited and empowered by massive training while enjoying potentially efficient inference. However, most existing NCO methods still struggle with real-time scaling-up performance; they are unable to solve routing problems involving thousands or tens of thousands of nodes in seconds, falling short of the needs of modern industries \cite{hou2023_tam}.

In answer to that, this work proposes GLOP (\textbf{G}lobal and \textbf{L}ocal \textbf{O}ptimization \textbf{P}olicies) which partitions a large routing problem into sub-Travelling Salesman Problems (TSPs) and further partitions potentially large (sub-)TSPs into small Shortest Hamiltonian Path Problems (SHPPs). GLOP hybridizes non-autoregressive (NAR) global partition and autoregressive (AR) local construction policies, where the global policy learns the first partition and the local policy learns to solve SHPPs. We intend to integrate the strengths while circumventing the drawbacks of NAR and AR paradigms (further discussed in Appendix
\ifincludeappendix \ref{app: discussions of two paradigms}).
\else B).
\fi
In particular, partitioning nodes into subsets (each corresponding to a TSP) well suits NAR heuristics, because it is a large-scale but coarse-grained task agnostic of within-subset node ordering. On the other hand, solving SHPPs could be efficiently handled with the AR heuristic because it is a small-scale but fine-grained task.

The solution pipeline of GLOP is applicable to variations of routing problems, such as those tackled in \cite{li2021hcvrp, zhang2022mtsptwr, zhang2021dac_vrp1, alesiani2022dac_vrp2, miranda2017dac_vrp4, li2021csp}. We evaluate GLOP on canonical TSP, Asymmetric TSP (ATSP), Capacitated Vehicle Routing Problem (CVRP), and Prize Collecting TSP (PCTSP). GLOP for (A)TSP, as opposed to most methods that require scale-specific and distribution-specific training, can perform consistently and competitively across scales, across distributions, and on real-world benchmarks, using the same set of local policies. Notably, it is the first neural solver to effectively scale to TSP100K, obtaining a 5.1\% optimality gap and a $174\times$ speedup compared with 1-run 1-trial LKH-3. GLOP for CVRP clearly outperforms prior state-of-the-art (SOTA) real-time solvers \cite{hou2023_tam} while using $10\times$ less execution time. On PCTSP, GLOP surpasses both recent neural solvers and conventional solvers. 

Accordingly, we summarize our contributions as follows: 
\begin{itemize}
    \item We propose GLOP, a versatile framework that extends existing neural solvers to large-scale problems. To our knowledge, it makes the first effective attempt at hybridizing NAR and AR end-to-end NCO paradigms.
    \item We propose to learn global partition heatmaps for decomposing large-scale routing problems, leveraging NAR heatmap learning in a novel way.
    \item We propose a one-size-fits-all real-time (A)TSP solver that learns small SHPP solution construction for arbitrarily large (A)TSP. We dispense with learning upper-level TSP policies suggested in \cite{LCP, pan2023htsp} while achieving better performance.
    \item On (A)TSP, GLOP delivers competitive scaling-up and cross-distribution performance and is the first neural solver to scale to TSP100K effectively. On CVRP and PCTSP, GLOP achieves SOTA real-time performance. 
\end{itemize}

\section{Background and related work}\label{sec: related work}
\subsection{Neural Combinatorial Optimization (NCO)}
Recent advances in NCO show promise for solving combinatorial optimization problems in an end-to-end manner \cite{survey1, survey2, berto2023rl4co}. The end-to-end neural routing solvers can be categorized into two paradigms: AR solution construction and NAR heatmap generation coupled with subsequent decoding.
\ifincludeappendix We defer further discussions to Appendix \ref{app: additional related work}.
\else We defer further discussions to Appendix C.
\fi

\subsection{Divide and conquer for VRP}
The idea of ``divide and conquer'' has long been applied to VRP variants in traditional (meta) heuristics \cite{zhang2021dac_vrp1, alesiani2022dac_vrp2, xiao2019dac_vrp3, taillard2019popmusic4tsp}. Recently, such an idea has been introduced in neural routing solvers. \citet{delegate2021} propose learning to delegate (L2D) the improvement of subtours to LKH-3 \cite{LKH3}. \citet{zong2022rbg} introduce Rewriting-by-Generating (RBG) framework that involves repeated learning-based merging and rule-based decomposition. However, both methods rely on iterative refinement, therefore holding back the real-time performance. More related to GLOP, \citet{hou2023_tam} present a Two-stage Divide Method (TAM), the prior SOTA real-time neural solver for large-scale CVRP, where a dividing model learns to partition CVRP into sub-TSPs autoregressively, and the sub-TSPs are then solved by sub-solvers such as Attention Model (AM) \cite{Attention2018} or LKH-3. Unlike other prior works, both TAM and GLOP target large-scale CVRP under real-time settings. By comparison, GLOP outperforms TAM on CVRP by leveraging more effective global representations and better neural sub-TSP solvers, and can also handle routing problems that TAM does not address.

\subsection{Local construction for TSP}

Learning local subtour reconstruction for TSP is initially introduced by \citet{LCP} in Learning Collaborative Policy (LCP). LCP generates diversified initial solutions (seeds) with neural models (seeders), then repeatedly decomposes and reconstructs them. However, LCP, limited mainly by the design of seeders, can hardly scale up to TSP with hundreds of nodes. More recently, \citet{pan2023htsp} propose H-TSP, a hierarchical TSP solver interleaving forming open-loop TSP (a.k.a., SHPP) with upper-level policies and conquering it. By comparison, GLOP dispenses with learning any upper-level TSP policy but is able to outperform H-TSP. Another concurrent work, namely select-and-optimize (SO) \cite{cheng23_select_and_optimize}, utilizes a TSP solution pipeline similar to GLOP. But SO heavily relies on sophisticated heuristics specific to TSP, resulting in prolonged computational time. By comparison, GLOP achieves competitive solutions while being hundreds of times more efficient.

\begin{figure*}[!t]
\includegraphics[width=\textwidth]{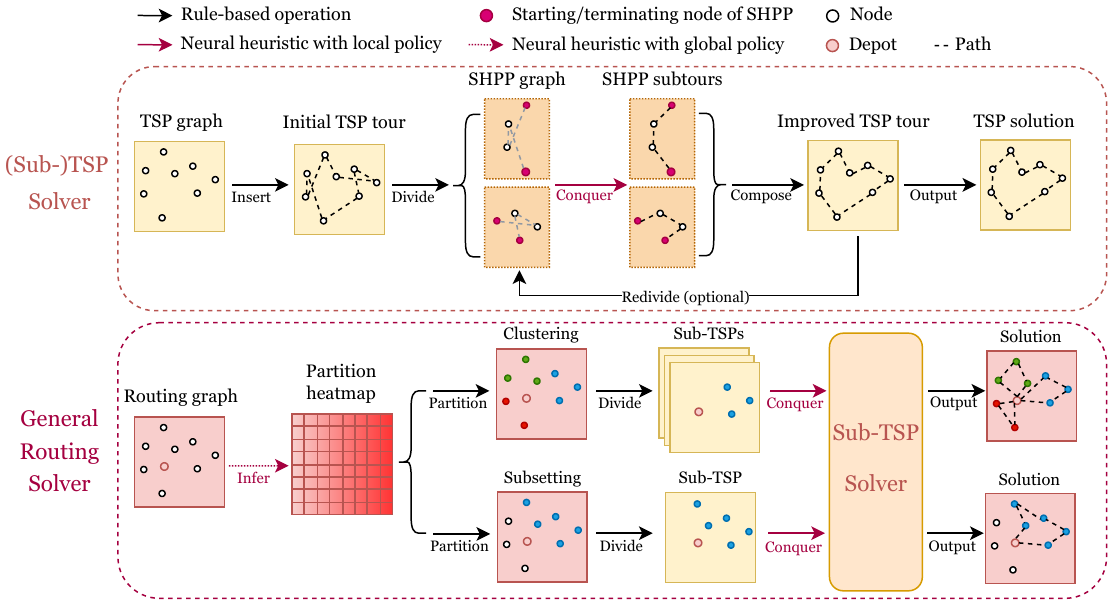}
\caption{The pipeline of GLOP.}
\label{fig: pipeline}
\end{figure*}

\section{Methodology overview}
GLOP is schematically illustrated in Figure \ref{fig: pipeline}. It aims to provide a unified and scalable framework for heterogeneous vehicle routing problems. To this end, our design targets three representative problem settings: (1) large-scale TSP alone, (2) large-scale CVRP requiring problem partitioning and solving multiple small sub-TSPs, and (3) large-scale PCTSP requiring problem partitioning and solving a single large sub-TSP.
\ifincludeappendix We defer the detailed explanations of these problems to Appendix \ref{app: problem definitions}.
\else We defer the detailed explanations of these problems to Appendix D.
\fi

In general, GLOP learns local policies for (sub-)TSP and global policies for partitioning general routing problems into sub-TSPs. Our (sub-)TSP solver generates initial TSP tours using Random Insertion, divides the complete tours into independent subtours, and learns to reconstruct them for improvements. Our general routing solver additionally learns to perform node clustering or subsetting that generates sub-TSP(s). We elaborate on our local policy and global policy in Section \ref{sec: GLOP for (sub) TSP} and Section \ref{sec: GLOP for CVRP/PCTSP}, respectively, 
\ifincludeappendix
and provide more details in Appendix \ref{app: Details of GLOP}.
\else
and provide more details in Appendix A.
\fi

\section{(Sub-)TSP solver}\label{sec: GLOP for (sub) TSP}
\subsection{Inference pipeline}

GLOP learns local policies to improve a TSP solution by decomposing and reconstructing it.

\paragraph{Initialization}
GLOP generates an initial TSP tour with Random Insertion (RI), a simple and generic heuristic. RI greedily picks the insertion place that minimizes the insertion cost for each node. 

\paragraph{}Then, GLOP performs improvements on the initial tour. Following \citet{LCP}, we refer to a round of improvement as a ``revision''; we refer to a local policy parameterized by an autoregressive NN and trained to solve SHPP$n$ (SHPP of $n$ nodes) as ``Reviser-$n$''. A revision involves decomposing and reconstructing the initial tour, comprising four sequential steps outlined below.

\paragraph{Decomposition}
When improved by Reviser-$n$, a complete tour with $N$ nodes is randomly decomposed into $\lfloor \frac{N}{n} \rfloor$ subtours, each with $n$ nodes. There is no overlap between every two subtours. A ``tail subtour'' with $N \bmod n$ nodes, if any, is left untouched until composition. Each subtour corresponds to an SHPP graph, and reconstructing a subtour is equivalent to solving an SHPP instance. We pick the decomposition positions uniformly when performing repeated revisions.

\paragraph{Transformation and augmentation}
To improve the predictability and homogeneity of the model inputs, we apply Min-max Normalization and an optional rotation to the SHPP graphs. They scale the x-axis coordinates to the range $[0,1]$ and set the lower bound of the y-axis to 0. In addition, we augment the SHPP instances by flipping the node coordinates to enhance the model performance.

\paragraph{Solving SHPPs with local policies}
We autoregressively reconstruct the subtours (i.e., solve the SHPP instances) with trainable revisers. Any SHPP solutions that are worse than the current ones will be discarded. This key step is detailed in Section \ref{sec: Solving SHPP with local policy}.

\paragraph{Composition}
The $\lfloor \frac{N}{n} \rfloor$ reconstructed (or original) subtours and a tail subtour, if any, compose an improved complete tour by connecting the starting/terminating nodes of SHPPs in their original order.

\paragraph{}GLOP can apply multiple revisers to solve a problem from different angles. Also, a single reviser can decompose the tour at different points and repeat its revisions. After all revisions, GLOP outputs the improved tour as its final solution. Notably, GLOP allows applying a single set of small-SHPP-trained models for arbitrarily large TSPs.


\subsection{Solving SHPP with local policy}\label{sec: Solving SHPP with local policy}

\paragraph{Problem formulation and motivation}
SHPP is also referred to as open-loop TSP. With starting/terminating nodes fixed, it aims to minimize the length of a Hamiltonian path visiting all nodes in between exactly once. Solving small SHPPs with neural networks, instead of directly solving TSP or with traditional heuristics, makes GLOP a highly parallelizable one-size-fits-all solution.


\paragraph{Model}
We parameterize our local policies based on Attention Model (AM) \cite{Attention2018}. To apply it to SHPP, we adjust its context embedding following \citet{LCP} and leverage the solution symmetries by autoregressively constructing solutions from both starting/determining nodes.

\paragraph{Local policy}
Given an SHPP instance $\bm{s}$ with starting/terminating node 1 and $n$, our stochastic local policy $p_{\bm{\theta}}( \bm{\omega}_{fd}, \bm{\omega}_{bd} \mid \bm{s})$, parameterized by the neural model $\bm{\theta}$, denotes the conditional probability of constructing forward and backward-decoded solutions $\bm{\omega}_{fd}$ and $\bm{\omega}_{bd}$. We let $\bm{\omega}_{1: t-1}$ denote the partial solution at time step $t$, then the local policy can be factorized into probability distribution of per-step construction:
\begin{equation}
\begin{aligned}
    p_{\bm{\theta}}( \bm{\omega}_{fd}, \bm{\omega}_{bd} \mid \bm{s}) 
    =  
    p_{\bm{\theta}}( \bm{\omega}_{fd} \mid \bm{s}) \times  p_{\bm{\theta}}( \bm{\omega}_{bd} \mid \bm{s})
    \\
    =
    \prod_{t=1}^{n-2} p_{{\bm{\theta}}}(\omega_{t} \mid \bm{s}, \bm{\omega}_{1: t-1}, n)
    \times
    p_{{\bm{\theta}}}(\omega_{t} \mid \bm{s}, \bm{\omega}_{1: t-1}, 1).
\end{aligned}
\end{equation}
We accept the better one between $\bm{\omega}_{fd}$ and $\bm{\omega}_{bd}$ during inference while making use of both for training.

\subsection{Training algorithm}\label{sec: Bidirectional REINFORCE}
We train our parameterized local policy, i.e., a reviser, by minimizing the expected length of its constructed SHPP solutions:
\begin{equation}
\begin{aligned}
    \mbox{minimize} \quad \mathcal{L}(\bm{\theta} \mid \bm{s}) = \mathbb{E}_{\bm{\omega}_{fd}, \bm{\omega}_{bd} \sim p_{\bm{\theta}}(\bm{\omega}_{fd}, \bm{\omega}_{bd} \mid \bm{s})}
    \\
    [f_{SHPP}(\bm{\omega}_{fd}, \bm{s})+f_{SHPP}(\bm{\omega}_{bd}, \bm{s})],
\end{aligned}
\end{equation}
where $f_{SHPP}$ maps an SHPP solution to its length. We apply the REINFORCE-based gradient estimator \cite{williams1992REINFORCE} using the average path length of two greedy rollouts as a baseline. This training algorithm doubles the experience learned on each instance and enables a more reliable baseline by weighing the greedy rollouts of both directions. 

\subsubsection{Two-stage curriculum learning}

According to our coordinate transformation, we design a two-stage curriculum to improve the homogeneity and consistency between training and inference instances. We are motivated by the following observation: the inputs to revisers are 1) SHPP graphs with y-axis upper bounds ranging from 0 to 1 after our coordinate transformation, and also 2) the outputs of its preceding module. Therefore, stage 1 in our curriculum trains revisers using multi-distribution SHPPs with varied y-axis upper bounds, and stage 2 collaboratively fine-tunes all revisers following the inference pipeline.




\section{General routing solver}\label{sec: GLOP for CVRP/PCTSP}
Many routing problems can be formulated hierarchically, which requires node clustering (e.g. CVRP, mTSP, Capacitated Arc Routing Problem) or node subsetting (e.g. PCTSP, Orienteering Problem, Covering Salesman Problem), followed by solving multiple sub-TSPs or a single sub-TSP, respectively \cite{li2021hcvrp, zhang2022mtsptwr, zhang2021dac_vrp1, alesiani2022dac_vrp2, xiao2019dac_vrp3, li2021csp}. For these general routing problems, GLOP involves an additional global partition policy defined by a parameterized partition heatmap (Section \ref{sec: Parameterized partition heatmap}) and trained with parallel on-policy sampling without costly step-by-step neural decoding (Section \ref{sec: training heatmap}). The applicability of our global policy is also discussed (Section \ref{sec: applicability}).

\subsection{Global policy as partition heatmap}\label{sec: Parameterized partition heatmap}

\paragraph{Partition heatmap}
We introduce a parameterized partition heatmap $\bm{\mathcal{H}}_{\bm{\phi}}(\bm{\rho}) = [h_{ij}(\bm{\rho})]_{(n+1)\times (n+1)}$ where $\bm{\rho}$ is the input instance with $n+1$ nodes including node 0 as the depot. $h_{ij} \in \mathbb{R}^{+}$ represents the unnormalized probability of nodes $i$ and $j$ belonging to the same subset.

\paragraph{Model and input graph}
The partition heatmap is parameterized by an isomorphic GNN $\bm{\phi}$ \cite{joshi2019GCN, dimes}. Inputs to the model are sparsified graphs with features designed separately for different problems.
\ifincludeappendix We defer the full details to Appendix \ref{app: Inputs to global policy}.
\else We defer the full details to Appendix A.4.
\fi

\paragraph{Global policy}
For node clustering, GLOP partitions all nodes into multiple subsets, each corresponding to a sub-TSP to solve. For node subsetting, GLOP partitions all nodes into two subsets, i.e., the to-visit subset and the others, where the to-visit subset forms a sub-TSP to solve. Let $\bm{\pi}=\{\bm{\pi}^r\}_{r=1}^{|\bm{\pi}|}$ denote a complete partition and $\bm{\pi}^r=\{\pi_t^r\}_{t=1}^{|\bm{\pi}^r|}$ the $r$-th subset containing both regular nodes and the depot. Each subset begins and terminates at the depot; that is, $\pi_1^r = \pi_{|\bm{\pi}^r|}^r=0$. Given $\bm{\mathcal{H}}_{\bm{\phi}}(\bm{\rho})$, our global policy partitions all nodes into $|\bm{\pi}|$ subsets by sequentially sampling nodes while satisfying problem-specific constraints $\Theta$:
\begin{equation}
    p_{\bm{\phi}}
    (\bm{\pi} | \bm{\rho})
    =
    \left\{\begin{array}{ll}
    \displaystyle
    \prod\limits_{r=1}^{|\bm{\pi}|}
    \prod\limits_{t=1}^{|\bm{\pi}^r| -1}
    \frac{h_{\pi_t^r,\pi_{t+1}^r}(\bm{\rho})}
    {
    \sum_{k\in \mathcal{N}(\bm{\pi}^p)
    }
    h_{\pi_t^r, k}(\bm{\rho})
    },\text{if } \bm{\pi} \in \Theta,   \\
    0, \qquad \qquad \qquad \qquad \qquad \qquad \; \text{otherwise},
\end{array}\right.
\label{eq: sample global solution}
\end{equation}
where $\mathcal{N}(\bm{\pi}^p)$ is the set of feasible actions given the current partial partition. For our benchmark problems,
\ifincludeappendix the applied constraints $\Theta$ are given in Appendix \ref{app: details of our global policy}.
\else the applied constraints $\Theta$ are given in Appendix A.4.
\fi

\paragraph{Decoding}
We apply greedy (GLOP-G) and sampling (GLOP-S) heatmap decoding \cite{dimes} to draw the node partitions following our global policy.

\begin{table*}[!h]
\centering
\resizebox{\textwidth}{!}{
\begin{tabular}{l|ccc|ccc|ccc}
\toprule
\multicolumn{1}{c|}{\multirow{2}{*}{Method}} &  \multicolumn{3}{c|}{TSP500} & \multicolumn{3}{c|}{TSP1K} & \multicolumn{3}{c}{TSP10K} \\
\multicolumn{1}{c|}{} & Obj. & Gap(\%) & Time & Obj. & Gap(\%) & Time & Obj. & Gap(\%) & Time \\ \midrule
Concorde \cite{concorde} &16.55 & 0.00 & 40.9m & 23.12 & 0.00 & 8.2h & - & - & - \\
LKH-3  \cite{LKH3} &16.55 & 0.00 & 5.5m & 23.12 & 0.00 & 24m & 71.77 & 0.00 & 13h\\
Random Insertion &18.59 & 12.3 & $<$1s & 26.12 & 13.0 & $<$1s & 81.84 & 14.0 & 5.7s \\ 
\midrule
AM \cite{Attention2018} &22.60 & 36.6 & 5.8m & 42.53 & 84.0 & 22m & 430 & 499 & 3.5m \\
LCP \cite{LCP} &20.82 & 25.8 & 29m & 36.34 & 57.2 & 34m & 357 & 397 & 4.3m\\

GCN+MCTS $\times 12$ \cite{fu2021generalize} &16.96 & 2.48 & 2.4m+33s & 23.86 & 3.20 & 4.9m+1.2m & 75.73 & 5.50 & 7.1m+6.0m \\ 
POMO-EAS \cite{eas2021} &24.04 & 45.3 & 1.0h & 47.79 & 107 & 8.6h & \multicolumn{3}{c}{OOM}\\
DIMES+S \cite{dimes} &19.06 & 15.0 & 2.0m & 26.96 & 16.1 & 2.4m & 86.25 & 20.0 & 3.1m   \\
DIMES+MCTS $\times 12$ \cite{dimes} &17.01 & 2.78 & 1.0m+2.1m &  23.86 & 3.20 & 2.6m+1.0m & 76.02 & 5.90 & 13.7m+20m  \\
Tspformer* \cite{yang2023memory-efficient-transformer} &17.57 & 5.97 & 3.1m & 27.02 & 16.9 & 5.0m & - & - & - \\
H-TSP \cite{pan2023htsp} & - & - & - & 24.65 & 6.62 & 47s &  77.75 & 7.32 & 48s \\ 
Pointerformer \cite{jin2023pointerformer} &17.14 & 3.56 & 1.0m & 24.80 & 7.30 & 6.5m & - & - & - \\
DeepACO \cite{ye2023deepaco} & 16.94 & 2.36 & 4.3m & 23.85 & 3.16 & 1.1h & - & - & - \\
\midrule
GLOP &17.07 & 3.14 & \textbf{19s} & 24.01 & 3.85 & \textbf{34s} & 75.62 & 5.36 & \textbf{32s} \\
GLOP (more revisions)  &\textbf{16.91} & \textbf{1.99} & 1.5m & \textbf{23.84} & \textbf{3.11} & 3.0m & \textbf{75.29} & \textbf{4.90} & 1.8m \\ 
\bottomrule
\end{tabular}}
\caption{Comparison results on 128 TSP500, 128 TSP1K, and 16 TSP10K. For all experiments on TSP, ``Time'' is the total runtime for solving all instances. If it has two terms, they correspond to the runtime of heatmap generation and MCTS, respectively. OOM: out of our graphics memory (24GB). *:  Results are drawn from the original literature with runtime proportionally adjusted (128/100) to match the size of our test datasets. See Appendix
\ifincludeappendix \ref{app: inference settings} and \ref{app: Implementation details}
\else A.6 and F
\fi
for full implementation details of GLOP and the baselines, respectively.}
\label{tab: tsp-res-m}
\end{table*}


\begin{table}
\centering
\begin{tabular}{l|ccc}
\toprule
\multicolumn{1}{c|}{\multirow{2}{*}{Method}} & \multicolumn{3}{c}{TSP100K} \\
\multicolumn{1}{c|}{} & Obj. & Gap(\%) & Time \\ \midrule
LKH-3 $\{T=1 \}$ &  226.4 & 0.00 & 8.1h \\
Random Insertion & 258.5 & 14.2 & 1.7m \\ 
\midrule
AM &  \multicolumn{3}{c}{OOM} \\
LCP & \multicolumn{3}{c}{OOM} \\
GCN+MCTS $\times 12$ & \multicolumn{3}{c}{OOM} \\ 
POMO-EAS & \multicolumn{3}{c}{OOM} \\
DIMES+MCTS $\times 12$ & \multicolumn{3}{c}{OOM} \\
DIMES+S & 286.1 & 26.4 & 2.0m\\
H-TSP & \multicolumn{3}{c}{OOM} \\
Pointerformer & \multicolumn{3}{c}{OOM} \\
\midrule
GLOP & 240.0 & 6.01 & \textbf{1.8m} \\
GLOP (more revisions) & \textbf{238.0} & \textbf{5.10} & 2.8m \\
\bottomrule
\end{tabular}
\caption{Comparison results on a TSP100K instance.}
\label{tab:res-l}
\end{table}


\begin{table}
\centering
\begin{tabular}{l|cc}
\toprule
\multicolumn{1}{c|}{Method} &  Avg. gap(\%)  & Time          \\ \midrule
LCP $\{M=1280\}$     & 99.9          & 3.6m          \\
DACT $\{T=1K\}$       & 865           & 50m           \\
GCN+MCTS $\times 1$ & 1.10    & 7.5m              \\
POMO-EAS $\{T=60\}$ & 18.8          & 20m           \\
DIMES+MCTS $\times 1$ & 2.21             &  7.4m             \\
AMDKD+EAS $\{T=100\}$ &  7.86*         &  48m            \\
Pointerformer & 6.04& 48s\\
\midrule
GLOP                          &  1.53  & \textbf{42s} \\ 
GLOP (more revisions)                         &  \textbf{0.69}   & 2.6m \\ 
\bottomrule
\multicolumn{3}{l}{*: Two instances are skipped due to OOM issue.}
\end{tabular}
\caption{Comparison results on TSPLIB instances.}
\label{tab:tsplib}
\end{table}

\begin{table}[]
\centering
\begin{tabular}{c|cc}
\toprule
Method   & MatNet \cite{kwon2021matnet} & GLOP        \\
\midrule
ATSP150  & 2.88 (7.2s) & \textbf{1.89} (8.2s) \\
ATSP250  & 4.49 (12s) & \textbf{2.04} (9.3s) \\
ATSP1000 & -           & \textbf{2.33} (15s) \\
\bottomrule
\end{tabular}
\caption{Comparison results on ATSP. The results were updated in July 2024.}
\label{tab: atsp}
\end{table}


\subsection{Training algorithm}\label{sec: training heatmap}
We train our global policy to output partitions that could lead to the best-performing final solutions after solving sub-TSPs. Given each instance $\bm{\rho}$, the training algorithm infers partition heatmap $\bm{\mathcal{H}}_{\bm{\phi}}(\bm{\rho})$, samples node partitions in parallel, feeds the sampled partitions into GLOP for sub-TSP solutions, and optimizes the expected final performance:
\begin{equation}
    \mbox{min} \; \mathcal{L}(\bm{\phi}|\bm{\rho}) =
    \mathbb{E}_{\bm{\pi} \sim p_{\bm{\phi}}
    (\bm{\pi} | \bm{\rho})} 
    [
    \sum_{r=1}^{|\bm{\pi}|}
    f_{TSP}(
    GLOP_{\bm{\theta}}(\bm{\pi}^r, \bm{\rho})
    )
    ],
\end{equation}
where $f_{TSP}$ is a mapping from a sub-TSP solution to its length, and $GLOP_{\bm{\theta}}$ generates sub-TSP solutions with well-trained local policies. We apply the REINFORCE algorithm \cite{williams1992REINFORCE} with the averaged reward of sampled solutions for the same instance as a baseline. The baseline is respectively computed for each instance within each training batch. GLOP for sub-TSP, i.e. $GLOP_{\bm{\theta}}$, enables efficient training of our global policy on large-scale problems due to its parallelizability and scalability.

\subsection{Applicability}\label{sec: applicability}
Many routing problems can be formulated hierarchically, involving node clustering and/or node subsetting, depending on the problem formulation. Node clustering is used when the problem requires formulating multiple routes that cover all nodes, while node subsetting is used when the problem requires formulating a single route that covers a subset of nodes. In some cases, a routing problem may require both subsetting and clustering. Our global policy offers a unified formulation for all these scenarios. Additionally, it can easily handle constraints via masking if they can be anticipated while constructing node subsets. To handle more complex constraints, one can assign a large negative value to the rewards of infeasible solutions or apply post-processing techniques before solution evaluation.

\begin{table*}[!t]
\centering
\resizebox{\textwidth}{!}{
\begin{tabular}{l|c|ccccccc}
\toprule
\multicolumn{1}{c|}{\multirow{2}{*}{Method}} &  \multirow{2}{*}{Time} & Uniform & \multicolumn{2}{c}{Expansion} & \multicolumn{2}{c}{Explosion} & \multicolumn{2}{c}{Implosion} \\
\multicolumn{1}{c|}{} &  & Gap(\%) & Gap(\%) & Det.(\%) & Gap(\%) & Det.(\%) & Gap(\%) & Det.(\%) \\ \midrule
AM \cite{Attention2018} & 0.5h & 2.310 & 17.97 & 678 & 3.817 & 65 & 2.431 & 5.2 \\ \midrule
AM+HAC \cite{hac} & \multirow{2}{*}{0.5h} & 2.484 & 3.997 & \textbf{61} & 3.084 & 24 & 2.595 & 4.5 \\
GLOP & & \textbf{0.091} &\textbf{0.166} & 82 & \textbf{0.066} & \textbf{-27} & \textbf{0.082} & \textbf{-9.9} \\ \midrule
AMDKD+EAS \cite{KD2022} &\multirow{2}{*}{2.0h} & 0.078 & 0.165 & 112 & 0.048 & -39 & 0.079 & 1.3 \\
GLOP (more revisions) & &\textbf{0.048} & \textbf{0.076} & \textbf{60} & \textbf{0.028} & \textbf{-41} & \textbf{0.044} & \textbf{-8.3} \\ \bottomrule
\end{tabular}}
\caption{Comparison results on the OoD datasets. $\mbox{Det} = \mbox{Gap}_{OoD} / \mbox{Gap}_{U} - 1$, where $\mbox{Gap}_{OoD}$ and $\mbox{Gap}_{U}$ are the optimality gaps on an OoD dataset and the Uniform dataset, respectively.}
\label{tab: exp-cross-distribution}
\end{table*}


\begin{table*}[]
\centering
\begin{tabular}{l|cccccccc}
\toprule
\multicolumn{1}{c|}{\multirow{2}{*}{Method}} &
  \multicolumn{2}{c}{CVRP1K} &
  \multicolumn{2}{c}{CVRP2K} &
  \multicolumn{2}{c}{CVRP5K} &
  \multicolumn{2}{c}{CVRP7K} \\ 
\multicolumn{1}{c|}{} &    Obj.          & Time (s)        & Obj.          & Time (s)        & Obj.           & Time (s)        & Obj.           & Time (s)        \\ \midrule
LKH-3 \cite{LKH3}  & 46.4          & 6.2          & 64.9          & 20           & 175.7          & 152          & 245.0          & 501          \\
\midrule
AM \cite{Attention2018}      & 61.4          & 0.6          & 114.4         & 1.9          & 257.1          & 12           & 354.3          & 26           \\
L2I \cite{L2I}           &  93.2          & 6.3          & 138.8         & 25           & -              & -            & -              & -            \\
NLNS \cite{hottung2019nlns} &  53.5 & 198 & - & - & - & - & - & -  \\
L2D \cite{delegate2021}  &46.3 & 1.5 & 65.2 & 38 & - & - & - & -  \\
RBG \cite{zong2022rbg}  &  74.0 & 13           & 137.6   & 42           & -     & -            & -     & -            \\
TAM-AM \cite{hou2023_tam}  & 50.1          & 0.8          & 74.3          & 2.2          & 172.2          & 12           & 233.4          & 26           \\
TAM-LKH3 \cite{hou2023_tam}  &  46.3          & 1.8          & 64.8          & 5.6          & 144.6          & 17           & 196.9          & 33           \\
TAM-HGS \cite{hou2023_tam}   & -             & -            & -             & -            & 142.8          & 30           & 193.6          & 52           \\
\midrule
GLOP-G    & 47.1          & \textbf{0.4} & 63.5          & \textbf{1.2} & 141.9          & \textbf{1.7} & 191.7          & \textbf{2.4} \\
GLOP-G (LKH-3)  & \textbf{45.9} & 1.1          & \textbf{63.0} & 1.5          & \textbf{140.6} & 4.0          & \textbf{191.2} & 5.8          \\ \bottomrule
\end{tabular}
\caption{Comparison results on large-scale CVRP following the settings in \cite{hou2023_tam}. ``Time'' corresponds to the per-instance runtime. GLOP-G (LKH-3) applies LKH-3 as its sub-TSP solver.}
\label{tab: cvrp main results}
\end{table*}

\begin{table*}[]
\centering
\begin{tabular}{c|c|cccccccc|cccc}
\toprule
\multirow{2}{*}{Instance} & \multirow{2}{*}{Scale} & \multicolumn{2}{c}{AM} & \multicolumn{2}{c}{TAM-AM} & \multicolumn{2}{c}{LKH-3} & \multicolumn{2}{c|}{TAM-LKH3} & \multicolumn{2}{c}{GLOP} & \multicolumn{2}{c}{GLOP-LKH3} \\
 &  & Gap & Time & Gap & Time & Gap & Time & Gap & Time & Gap & Time & Gap & Time \\ \midrule
LEUVEN1 & 3001 & 46.9 & 10s & 20.2 & 10s & 18.1 & 69s & 19.3 & 16s & \textbf{16.9} & \textbf{2s} & \textbf{16.6} & \textbf{8s} \\
LEUVEN2 & 4001 & 53.3 & 13s & 38.6 & 14s & 22.1 & 74s & 15.9 & 24s & 21.8 & \textbf{3s} & 21.1 & \textbf{3s} \\
ANTWERP1 & 6001 & 39.3 & 13s & 24.9 & 13s & 24.2 & 596s & 24.0 & 25s & \textbf{20.3} & \textbf{3s} & \textbf{19.3} & \textbf{14s} \\
ANTWERP2 & 7001 & 50.3 & 15s & 33.2 & 15s & 31.1 & 479s & 22.6 & 32s & \textbf{19.4} & \textbf{4s} & \textbf{19.4} & \textbf{7s} \\ \midrule
GHENT1 & 10001 & 46.9 & 21s & 30.2 & 22s & \multicolumn{2}{c}{-} & 29.5 & 37s & \textbf{20.3} & \textbf{5s} & \textbf{18.3} & \textbf{22s} \\
GHENT2 & 11001 & 52.2 & 39s & 33.3 & 38s & \multicolumn{2}{c}{-} & 23.7 & 56s & \textbf{19.8} & \textbf{6s} & \textbf{18.1} & \textbf{8s} \\
BRUSSELS1 & 15001 & 52.4 & 131s & 43.4 & 139s & \multicolumn{2}{c}{-} & 27.2 & 167s & 27.6 & \textbf{8s} & 27.5 & \textbf{26s} \\
BRUSSELS2 & 16001 & 52.4 & 166s & 39.0 & 159s & \multicolumn{2}{c}{-} & 37.1 & 187s & \textbf{22.4} & \textbf{9s} & \textbf{20.1} & \textbf{14s} \\
\bottomrule
\end{tabular}
\caption{Comparison results on large-scale CVRPLIB instances.}
\label{tab: cvrplib}
\end{table*}

\begin{table*}[]
\centering
\begin{tabular}{l|cccccc}
\toprule
\multicolumn{1}{c|}{\multirow{2}{*}{Method}} & \multicolumn{2}{c}{PCTSP500} & \multicolumn{2}{c}{PCTSP1K} & \multicolumn{2}{c}{PCTSP5K} \\  
\multicolumn{1}{c|}{}  & Obj. & Time & Obj. & Time & Obj. & Time \\ \midrule
OR Tools & 15.0 & 1h & 24.9 & 1h & 63.3 & 1h \\
OR Tools (more iterations) & 14.4 & 16h & 20.6 & 16h & 54.4 & 16h \\
\midrule
AM \cite{Attention2018} & 19.3 & 14m & 34.8 & 23m & 175 & 21m \\
MDAM \cite{MDAM} & 14.8 & 2.8m & 22.2 & 17m & 58.9 & 3h \\ \midrule
GLOP-G  & 14.6 & \textbf{26s} & 20.0 & \textbf{47s} & 46.0 & \textbf{3.7m} \\
GLOP-S  & \textbf{14.3} & 1.5m & \textbf{19.8} & 2.5m & \textbf{44.9} & 16m \\ \bottomrule
\end{tabular}
\caption{Comparison results of GLOP and the baselines on 128 PCTSP500, 1K, and 5K. ``Time'' corresponds to the total execution time for solving all instances.}
\label{tab: exp-pctsp}
\end{table*}

\section{Experimentation}\label{section: exp}
\subsection{Experimental Setup}
\paragraph{Datasets}
We refer the readers to \citet{Attention2018} for more specific definitions of benchmark problems. (1) We evaluate GLOP on uniformly sampled large-scale TSP instances (i.e., TSP500, 1K, and 10K) used in \citet{fu2021generalize} and an additionally generated TSP100K instance. We perform a cross-distribution evaluation on test instances used in \cite{KD2022}. For evaluation on real-world benchmarks, we draw all 49 symmetric TSP instances featuring \textit{EUC\_2D} and containing fewer than 1000 nodes (since most baselines cannot process larger-scale instances) from TSPLIB and map all instances to the $[0,1]^2$ square through Min-max Normalization. The test datasets of ATSP are generated following \citet{kwon2021matnet}. (2) For CVRP, we adhere to the settings in TAM \cite{hou2023_tam} and use the code of AM \cite{Attention2018} to generate test datasets on CVRP1K, 2K, 5K, and 7K, each containing 100 instances. We also evaluate GLOP on several large-scale CVRPLIB instances. (3) For PCTSP, we follow the settings in AM \cite{Attention2018} for data generation on PCTSP500, 1K, and 5K. As suggested by \citet{Attention2018}, we specify $K^n = 9, 12, 20$ to sample prizes for $n=$500, 1K, 5K, respectively , i.e., $\beta_i \sim \mbox{Uniform}(0, 3\frac{K^n}{n})$.

\paragraph{Baselines}
Evaluating an NCO method typically involves two metrics: the objective value and the runtime. To ensure the validity of our comparisons, we select SOTA baselines with adjustable runtime that can match GLOP.
\ifincludeappendix We defer detailed implementations of the baselines to Appendix \ref{app: Implementation details}.
\else We defer detailed implementations of the baselines to Appendix F.
\fi

\paragraph{Hardware}
Unless otherwise stated, GLOP and the baselines are executed on a 12-core Intel(R) Xeon(R) Platinum 8255C CPU and an NVIDIA RTX 3090 Graphics Card.

\subsection{Travelling Salesman Problem}

\paragraph{Large-scale TSP}
Comparison results on large-scale TSP are shown in Table \ref{tab: tsp-res-m} and Table \ref{tab:res-l}. For GCN+MCTS \cite{fu2021generalize} and DIMES+MCTS \cite{dimes}, we use all 12 CPU cores for MCTS and limit its running time to ensure comparable results. Concluded from the results, GLOP is highly efficient due to its decomposed solution scheme
\ifincludeappendix (see Appendix \ref{app: discussing time complexity} for the analysis of time complexity).
\else (see Appendix E.5 for the analysis of time complexity).
\fi
Furthermore, the memory consumption of GLOP can be basically invariant of the problem scale if reconstructing the subtours using a fixed batch size. Hence, it is the first neural solver to effectively scale to TSP100K, obtaining a 5.1\% optimality gap and a $174\times$ speed-up compared to LKH-3. Compared with LCP \cite{LCP} and H-TSP \cite{pan2023htsp}, GLOP dispenses with learning upper-level TSP policies while achieving better performance. Compared with NAR methods conducting MCTS refinements \cite{fu2021generalize, dimes}, GLOP generates reasonable solutions even before they have finished initialization or produced prerequisite heatmaps for solution decoding. Hence, GLOP exhibits clear advantages for real-time applications.

\paragraph{Cross-distribution TSP}
Table \ref{tab: exp-cross-distribution} gathers the comparison between GLOP and two baselines specially devised for cross-distribution performance \cite{hac, KD2022} on four TSP100 datasets with different distributions, i.e., uniform, expansion, explosion, and implosion. Recall that we use uniformly distributed samples to train GLOP. Hence, the latter three datasets contain out-of-distribution (OoD) instances. Results show that GLOP obtains smaller gaps and less Det. on most OoD datasets. We argue that the holistic solution scheme of GLOP is the main contributor to its cross-distribution performance. The inputs to the neural models are local SHPP graphs, making GLOP insensitive to the overall TSP distribution.
\ifincludeappendix We conduct further discussions in Appendix \ref{app: cd contributer}.
\else We conduct further discussions in Appendix E.3.
\fi

\paragraph{Real-world TSPLIB}
We evaluate GLOP and the baselines on real-world TSPLIB instances and collect the results in Table \ref{tab:tsplib}, where GLOP performs favorably against the baselines due to its consistent performance across scales and distributions. Note that we test each instance individually without parallel computation.

\paragraph{Asymmetric TSP\footnote{We have updated the ATSP results in the new version of the paper (July 2024). After fixing a bug, we achieved better performance. For more details, see \url{https://github.com/henry-yeh/GLOP/pull/3}.}}

GLOP is compatible with any neural architecture and can be extended to asymmetric distance. Table \ref{tab: atsp} exemplifies this flexibility on ATSP, where we replace AM with MatNet \cite{kwon2021matnet} which is specially designed for ATSP.
We follow the experimental setup in \cite{kwon2021matnet} and generalize MatNet100 as baseline. For GLOP, we apply MatNet50 checkpoint as our reviser without retraining. The results validate that GLOP can successfully extend MatNet to solve large ATSP instances. Note that MatNet is limited to problem scales no larger than 256 due to its one-hot initialization while there is no such limitation for GLOP-empowered MatNet.

\subsection{Capacitated Vehicle Routing Problem}
GLOP for CVRP involves node clustering with our global policy, followed by solving the produced sub-TSPs with our sub-TSP solver.

\paragraph{Large-scale CVRP}
Table \ref{tab: cvrp main results} summarizes the results of the comparison in large-scale CVRP, where the RBG \cite{zong2022rbg} and L2D \cite{delegate2021} models are generalized for evaluation here due to their different training settings, and the results of other baselines are drawn from \cite{hou2023_tam}. Here, we apply the global policy trained on CVRP2K to both CVRP5K and CVRP7K, verifying the generalization performance of GLOP. Compared to the methods that entail iterative solution refinement \cite{LKH3, delegate2021, L2I, zong2022rbg}, both TAM \cite{hou2023_tam} and GLOP can deliver more real-time solutions. Compared with prior SOTA real-time solver TAM, GLOP learns more effective global/local policies and enables higher decoding efficiency. Hence, GLOP outperforms TAM regarding both solution quality and efficiency.

\paragraph{Real-world CVRPLIB}
We test GLOP on large-scale CVRPLIB instances and present results in Table \ref{tab: cvrplib}. We generalize the model trained on CVRP2000 to these large instances and compare GLOP to prior SOTA real-time solver TAM \cite{hou2023_tam}. The results of TAM and other baselines are drawn from \cite{hou2023_tam}. The comparison also demonstrates the superiority of GLOP in both solution quality and efficiency, especially for very large instances.

\subsection{Prize Collecting Travelling Salesman Problem}
GLOP for PCTSP involves node subsetting with our global policies, followed by solving a sub-TSP instance with our sub-TSP solvers. Large-scale PCTSP entails both a scalable global policy to generate a promising node partition and a scalable local policy to tackle the equivalently large sub-TSP. We evaluate greedy (GLOP-G) and sampling (GLOP-S) decoding for our global policy. The comparison results on large-scale PCTSP are displayed in Table \ref{tab: exp-pctsp}, where GLOP surpasses recent neural solvers and conventional solvers in terms of both solution quality and efficiency. 

\section{Conclusion and limitation}
This paper proposes GLOP to learn global policies for coarse-grained problem partitioning and local policies for fine-grained route construction. GLOP leverages the scalability of the NAR paradigm and meticulousness of the AR paradigm, making the first effective attempt at hybridizing them. Extensive evaluations on large-scale TSP, ATSP, CVRP, and PCTSP demonstrate its competitive and SOTA real-time performance.
However, GLOP might be less competitive in application scenarios where prolonged execution time is allowed. In terms of its ability to trade off execution time for solution quality, GLOP might be inferior to the methods based on iterative solution refinement (further discussed in Appendix E.1). Our future focus will be on addressing this limitation.
In addition, we plan to investigate the emerging possibilities that arise when viewing AR and NAR methods from a unified perspective. We believe it is also promising to exploit unsupervised Deep Graph Clustering techniques \cite{yue2022deep-graph-clustering-survey, liuyue_RGC} or to formulate node classification tasks to solve large-scale routing problems hierarchically.

\section*{Acknowledgments}
The authors appreciate the helpful discussions with Juntao Li, Yu Hong, and anonymous reviewers. Yu Hu, Jiusi Yin, and Tao Yu also contributed to this work. This research was supported by the National Natural Science Foundation of China (NSFC) [grant number 61902269, 62176172]; the National Key R\&D Program of China [grant number 2018YFA0701700, 2018YFA0701701]; the Singapore Ministry of Education (MOE) Academic Research Fund (AcRF) Tier 1 grant; the Undergraduate Training Program for Innovation and Entrepreneurship, Soochow University [grant number 202210285001Z, 202310285041Z]; the Priority Academic Program Development of Jiangsu Higher Education Institutions, China; and Provincial Key Laboratory for Computer Information Processing Technology, Soochow University [grant number KJS1938].

\bibliography{main}

\begin{thebibliography}{72}
\providecommand{\natexlab}[1]{#1}

\bibitem[{Alesiani, Ermis, and Gkiotsalitis(2022)}]{alesiani2022dac_vrp2}
Alesiani, F.; Ermis, G.; and Gkiotsalitis, K. 2022.
\newblock Constrained Clustering for the Capacitated Vehicle Routing Problem (CC-CVRP).
\newblock \emph{Applied artificial intelligence}, 36(1): 1995658.

\bibitem[{Applegate et~al.(2006)Applegate, Bixby, Chvatal, and Cook}]{concorde}
Applegate, D.; Bixby, R.; Chvatal, V.; and Cook, W. 2006.
\newblock Concorde TSP solver.

\bibitem[{Bello et~al.(2016)Bello, Pham, Le, Norouzi, and Bengio}]{belloRL}
Bello, I.; Pham, H.; Le, Q.~V.; Norouzi, M.; and Bengio, S. 2016.
\newblock Neural combinatorial optimization with reinforcement learning.
\newblock \emph{ArXiv preprint}, abs/1611.09940.

\bibitem[{Berto et~al.(2023)Berto, Hua, Park, Kim, Kim, Son, Kim, Kim, and Park}]{berto2023rl4co}
Berto, F.; Hua, C.; Park, J.; Kim, M.; Kim, H.; Son, J.; Kim, H.; Kim, J.; and Park, J. 2023.
\newblock Rl4co: an extensive reinforcement learning for combinatorial optimization benchmark.
\newblock \emph{arXiv preprint arXiv:2306.17100}.

\bibitem[{Bi et~al.(2022)Bi, Ma, Wang, Cao, Chen, Sun, and Chee}]{KD2022}
Bi, J.; Ma, Y.; Wang, J.; Cao, Z.; Chen, J.; Sun, Y.; and Chee, Y.~M. 2022.
\newblock Learning Generalizable Models for Vehicle Routing Problems via Knowledge Distillation.
\newblock \emph{ArXiv preprint}, abs/2210.07686.

\bibitem[{Bogyrbayeva et~al.(2022)Bogyrbayeva, Meraliyev, Mustakhov, and Dauletbayev}]{survey1}
Bogyrbayeva, A.; Meraliyev, M.; Mustakhov, T.; and Dauletbayev, B. 2022.
\newblock Learning to Solve Vehicle Routing Problems: A Survey.
\newblock \emph{ArXiv preprint}, abs/2205.02453.

\bibitem[{Bresson and Laurent(2021)}]{bresson2021transformer_for_tsp}
Bresson, X.; and Laurent, T. 2021.
\newblock The transformer network for the traveling salesman problem.
\newblock \emph{ArXiv preprint}, abs/2103.03012.

\bibitem[{Chen et~al.(2022)Chen, Zong, Zhuang, Yan, Jin, and Li}]{chen2022_mixed_delivery_pickup}
Chen, J.; Zong, Z.; Zhuang, Y.; Yan, H.; Jin, D.; and Li, Y. 2022.
\newblock Reinforcement Learning for Practical Express Systems with Mixed Deliveries and Pickups.
\newblock \emph{ACM Transactions on Knowledge Discovery from Data (TKDD)}.

\bibitem[{Cheng et~al.(2023)Cheng, Zheng, Cong, Jiang, and Pu}]{cheng23_select_and_optimize}
Cheng, H.; Zheng, H.; Cong, Y.; Jiang, W.; and Pu, S. 2023.
\newblock Select and Optimize: Learning to solve large-scale TSP instances.
\newblock In Ruiz, F.; Dy, J.; and van~de Meent, J.-W., eds., \emph{Proceedings of The 26th International Conference on Artificial Intelligence and Statistics}, volume 206 of \emph{Proceedings of Machine Learning Research}, 1219--1231. PMLR.

\bibitem[{Choo et~al.(2022)Choo, Kwon, Kim, Jae, Hottung, Tierney, and Gwon}]{sgbs}
Choo, J.; Kwon, Y.-D.; Kim, J.; Jae, J.; Hottung, A.; Tierney, K.; and Gwon, Y. 2022.
\newblock Simulation-guided beam search for neural combinatorial optimization.
\newblock \emph{ArXiv preprint}, abs/2207.06190.

\bibitem[{d~O~Costa et~al.(2020)d~O~Costa, Rhuggenaath, Zhang, and Akcay}]{2opt}
d~O~Costa, P.~R.; Rhuggenaath, J.; Zhang, Y.; and Akcay, A. 2020.
\newblock Learning 2-opt heuristics for the traveling salesman problem via deep reinforcement learning.
\newblock In \emph{Asian Conference on Machine Learning}, 465--480. PMLR.

\bibitem[{Fan et~al.(2022)Fan, Wu, Liao, Cao, Guo, Sartoretti, and Wu}]{fan2022UAV}
Fan, M.; Wu, Y.; Liao, T.; Cao, Z.; Guo, H.; Sartoretti, G.; and Wu, G. 2022.
\newblock Deep Reinforcement Learning for UAV Routing in the Presence of Multiple Charging Stations.
\newblock \emph{IEEE Transactions on Vehicular Technology}.

\bibitem[{Fu, Qiu, and Zha(2021)}]{fu2021generalize}
Fu, Z.; Qiu, K.; and Zha, H. 2021.
\newblock Generalize a Small Pre-trained Model to Arbitrarily Large {TSP} Instances.
\newblock In \emph{Thirty-Fifth {AAAI} Conference on Artificial Intelligence, {AAAI} 2021}, 7474--7482. {AAAI} Press.

\bibitem[{Goh et~al.(2022)Goh, Lee, Bresson, Laurent, and Lim}]{goh2022combining_rl_optimal_transport}
Goh, Y.~L.; Lee, W.~S.; Bresson, X.; Laurent, T.; and Lim, N. 2022.
\newblock Combining reinforcement learning and optimal transport for the traveling salesman problem.
\newblock \emph{ArXiv preprint}, abs/2203.00903.

\bibitem[{Grinsztajn, Furelos-Blanco, and Barrett(2022)}]{grinsztajn2022poppy}
Grinsztajn, N.; Furelos-Blanco, D.; and Barrett, T.~D. 2022.
\newblock Population-Based Reinforcement Learning for Combinatorial Optimization.
\newblock \emph{ArXiv preprint}, abs/2210.03475.

\bibitem[{Helsgaun(2017)}]{LKH3}
Helsgaun, K. 2017.
\newblock An extension of the Lin-Kernighan-Helsgaun TSP solver for constrained traveling salesman and vehicle routing problems.
\newblock \emph{Roskilde: Roskilde University}, 24--50.

\bibitem[{Hottung, Kwon, and Tierney(2022)}]{eas2021}
Hottung, A.; Kwon, Y.; and Tierney, K. 2022.
\newblock Efficient Active Search for Combinatorial Optimization Problems.
\newblock In \emph{The Tenth International Conference on Learning Representations, {ICLR} 2022, Virtual Event, April 25-29, 2022}. OpenReview.net.

\bibitem[{Hottung and Tierney(2019)}]{hottung2019nlns}
Hottung, A.; and Tierney, K. 2019.
\newblock Neural large neighborhood search for the capacitated vehicle routing problem.
\newblock \emph{ArXiv preprint}, abs/1911.09539.

\bibitem[{Hou et~al.(2023)Hou, Yang, Su, Wang, and Deng}]{hou2023_tam}
Hou, Q.; Yang, J.; Su, Y.; Wang, X.; and Deng, Y. 2023.
\newblock Generalize Learned Heuristics to Solve Large-scale Vehicle Routing Problems in Real-time.
\newblock In \emph{The Eleventh International Conference on Learning Representations}.

\bibitem[{Hudson et~al.(2022)Hudson, Li, Malencia, and Prorok}]{hudson2021gls}
Hudson, B.; Li, Q.; Malencia, M.; and Prorok, A. 2022.
\newblock Graph Neural Network Guided Local Search for the Traveling Salesperson Problem.
\newblock In \emph{The Tenth International Conference on Learning Representations, {ICLR} 2022, Virtual Event, April 25-29, 2022}. OpenReview.net.

\bibitem[{Jiang et~al.(2023)Jiang, Cao, Wu, Song, and Zhang}]{jiang2023ensemble}
Jiang, Y.; Cao, Z.; Wu, Y.; Song, W.; and Zhang, J. 2023.
\newblock Ensemble-based Deep Reinforcement Learning for Vehicle Routing Problems under Distribution Shift.
\newblock In \emph{Thirty-seventh Conference on Neural Information Processing Systems}.

\bibitem[{Jiang et~al.(2022)Jiang, Wu, Cao, and Zhang}]{dro}
Jiang, Y.; Wu, Y.; Cao, Z.; and Zhang, J. 2022.
\newblock Learning to Solve Routing Problems via Distributionally Robust Optimization.
\newblock In \emph{Thirty-Sixth {AAAI} Conference on Artificial Intelligence, {AAAI} 2022}, 9786--9794. {AAAI} Press.

\bibitem[{Jin et~al.(2023)Jin, Ding, Pan, He, Zhao, Qin, Song, and Bian}]{jin2023pointerformer}
Jin, Y.; Ding, Y.; Pan, X.; He, K.; Zhao, L.; Qin, T.; Song, L.; and Bian, J. 2023.
\newblock Pointerformer: Deep Reinforced Multi-Pointer Transformer for the Traveling Salesman Problem.
\newblock \emph{ArXiv preprint}, abs/2304.09407.

\bibitem[{Joshi et~al.(2022)Joshi, Cappart, Rousseau, and Laurent}]{Generalization2022}
Joshi, C.~K.; Cappart, Q.; Rousseau, L.-M.; and Laurent, T. 2022.
\newblock Learning the travelling salesperson problem requires rethinking generalization.
\newblock \emph{Constraints}, 1--29.

\bibitem[{Joshi, Laurent, and Bresson(2019)}]{joshi2019GCN}
Joshi, C.~K.; Laurent, T.; and Bresson, X. 2019.
\newblock An efficient graph convolutional network technique for the travelling salesman problem.
\newblock \emph{ArXiv preprint}, abs/1906.01227.

\bibitem[{Kim, Park, and Kim(2021)}]{LCP}
Kim, M.; Park, J.; and Kim, J. 2021.
\newblock Learning Collaborative Policies to Solve NP-hard Routing Problems.
\newblock In Ranzato, M.; Beygelzimer, A.; Dauphin, Y.~N.; Liang, P.; and Vaughan, J.~W., eds., \emph{Advances in Neural Information Processing Systems 34: Annual Conference on Neural Information Processing Systems 2021, NeurIPS 2021, December 6-14, 2021, virtual}, 10418--10430.

\bibitem[{Kim, Park, and Park(2022)}]{symNCO}
Kim, M.; Park, J.; and Park, J. 2022.
\newblock Sym-NCO: Leveraging Symmetricity for Neural Combinatorial Optimization.
\newblock \emph{ArXiv preprint}, abs/2205.13209.

\bibitem[{Kingma and Ba(2015)}]{Adam}
Kingma, D.~P.; and Ba, J. 2015.
\newblock Adam: {A} Method for Stochastic Optimization.
\newblock In Bengio, Y.; and LeCun, Y., eds., \emph{3rd International Conference on Learning Representations, {ICLR} 2015, San Diego, CA, USA, May 7-9, 2015, Conference Track Proceedings}.

\bibitem[{Kool et~al.(2022)Kool, van Hoof, Gromicho, and Welling}]{kool2022ddp}
Kool, W.; van Hoof, H.; Gromicho, J.; and Welling, M. 2022.
\newblock Deep policy dynamic programming for vehicle routing problems.
\newblock In \emph{International Conference on Integration of Constraint Programming, Artificial Intelligence, and Operations Research}, 190--213. Springer.

\bibitem[{Kool, van Hoof, and Welling(2019)}]{Attention2018}
Kool, W.; van Hoof, H.; and Welling, M. 2019.
\newblock Attention, Learn to Solve Routing Problems!
\newblock In \emph{7th International Conference on Learning Representations, {ICLR} 2019, New Orleans, LA, USA, May 6-9, 2019}. OpenReview.net.

\bibitem[{Kwon et~al.(2020)Kwon, Choo, Kim, Yoon, Gwon, and Min}]{POMO}
Kwon, Y.; Choo, J.; Kim, B.; Yoon, I.; Gwon, Y.; and Min, S. 2020.
\newblock {POMO:} Policy Optimization with Multiple Optima for Reinforcement Learning.
\newblock In Larochelle, H.; Ranzato, M.; Hadsell, R.; Balcan, M.; and Lin, H., eds., \emph{Advances in Neural Information Processing Systems 33: Annual Conference on Neural Information Processing Systems 2020, NeurIPS 2020, December 6-12, 2020, virtual}.

\bibitem[{Kwon et~al.(2021)Kwon, Choo, Yoon, Park, Park, and Gwon}]{kwon2021matnet}
Kwon, Y.; Choo, J.; Yoon, I.; Park, M.; Park, D.; and Gwon, Y. 2021.
\newblock Matrix encoding networks for neural combinatorial optimization.
\newblock In Ranzato, M.; Beygelzimer, A.; Dauphin, Y.~N.; Liang, P.; and Vaughan, J.~W., eds., \emph{Advances in Neural Information Processing Systems 34: Annual Conference on Neural Information Processing Systems 2021, NeurIPS 2021, December 6-14, 2021, virtual}, 5138--5149.

\bibitem[{Li et~al.(2021{\natexlab{a}})Li, Ma, Gao, Cao, Lim, Song, and Zhang}]{li2021hcvrp}
Li, J.; Ma, Y.; Gao, R.; Cao, Z.; Lim, A.; Song, W.; and Zhang, J. 2021{\natexlab{a}}.
\newblock Deep reinforcement learning for solving the heterogeneous capacitated vehicle routing problem.
\newblock \emph{IEEE Transactions on Cybernetics}, 52(12): 13572--13585.

\bibitem[{Li et~al.(2021{\natexlab{b}})Li, Zhang, Wang, Wang, Han, and Wang}]{li2021csp}
Li, K.; Zhang, T.; Wang, R.; Wang, Y.; Han, Y.; and Wang, L. 2021{\natexlab{b}}.
\newblock Deep reinforcement learning for combinatorial optimization: Covering salesman problems.
\newblock \emph{IEEE transactions on cybernetics}, 52(12): 13142--13155.

\bibitem[{Li, Yan, and Wu(2021)}]{delegate2021}
Li, S.; Yan, Z.; and Wu, C. 2021.
\newblock Learning to delegate for large-scale vehicle routing.
\newblock In Ranzato, M.; Beygelzimer, A.; Dauphin, Y.~N.; Liang, P.; and Vaughan, J.~W., eds., \emph{Advances in Neural Information Processing Systems 34: Annual Conference on Neural Information Processing Systems 2021, NeurIPS 2021, December 6-14, 2021, virtual}, 26198--26211.

\bibitem[{Li, Chen, and Koltun(2018)}]{2018gcn_guided_tree_search}
Li, Z.; Chen, Q.; and Koltun, V. 2018.
\newblock Combinatorial Optimization with Graph Convolutional Networks and Guided Tree Search.
\newblock In Bengio, S.; Wallach, H.~M.; Larochelle, H.; Grauman, K.; Cesa{-}Bianchi, N.; and Garnett, R., eds., \emph{Advances in Neural Information Processing Systems 31: Annual Conference on Neural Information Processing Systems 2018, NeurIPS 2018, December 3-8, 2018, Montr{\'{e}}al, Canada}, 537--546.

\bibitem[{Liu et~al.(2023)Liu, Liang, Xia, Yang, Zhou, Liu, Liu, and Li}]{liuyue_RGC}
Liu, Y.; Liang, K.; Xia, J.; Yang, X.; Zhou, S.; Liu, M.; Liu, X.; and Li, S.~Z. 2023.
\newblock Reinforcement Graph Clustering with Unknown Cluster Number.
\newblock In \emph{Proceedings of the 31st ACM International Conference on Multimedia}, 3528--3537.

\bibitem[{Liu et~al.(2022)Liu, Xia, Zhou, Wang, Guo, Yang, Liang, Tu, Li, and Liu}]{yue2022deep-graph-clustering-survey}
Liu, Y.; Xia, J.; Zhou, S.; Wang, S.; Guo, X.; Yang, X.; Liang, K.; Tu, W.; Li, Z.~S.; and Liu, X. 2022.
\newblock A Survey of Deep Graph Clustering: Taxonomy, Challenge, and Application.
\newblock \emph{arXiv preprint arXiv:2211.12875}.

\bibitem[{Loshchilov and Hutter(2017)}]{loshchilov2016sgdr}
Loshchilov, I.; and Hutter, F. 2017.
\newblock {SGDR:} Stochastic Gradient Descent with Warm Restarts.
\newblock In \emph{5th International Conference on Learning Representations, {ICLR} 2017, Toulon, France, April 24-26, 2017, Conference Track Proceedings}. OpenReview.net.

\bibitem[{Loshchilov and Hutter(2019)}]{loshchilov2017AdamW}
Loshchilov, I.; and Hutter, F. 2019.
\newblock Decoupled Weight Decay Regularization.
\newblock In \emph{7th International Conference on Learning Representations, {ICLR} 2019, New Orleans, LA, USA, May 6-9, 2019}. OpenReview.net.

\bibitem[{Lu, Zhang, and Yang(2020)}]{L2I}
Lu, H.; Zhang, X.; and Yang, S. 2020.
\newblock A Learning-based Iterative Method for Solving Vehicle Routing Problems.
\newblock In \emph{8th International Conference on Learning Representations, {ICLR} 2020, Addis Ababa, Ethiopia, April 26-30, 2020}. OpenReview.net.

\bibitem[{Ma et~al.(2019)Ma, Ge, He, Thaker, and Drori}]{ma2019HRL}
Ma, Q.; Ge, S.; He, D.; Thaker, D.; and Drori, I. 2019.
\newblock Combinatorial optimization by graph pointer networks and hierarchical reinforcement learning.
\newblock \emph{ArXiv preprint}, abs/1911.04936.

\bibitem[{Ma, Cao, and Chee(2023)}]{ma2023neuopt}
Ma, Y.; Cao, Z.; and Chee, Y.~M. 2023.
\newblock Learning to Search Feasible and Infeasible Regions of Routing Problems with Flexible Neural k-Opt.
\newblock In \emph{Thirty-seventh Conference on Neural Information Processing Systems}.

\bibitem[{Ma et~al.(2021)Ma, Li, Cao, Song, Zhang, Chen, and Tang}]{DACT}
Ma, Y.; Li, J.; Cao, Z.; Song, W.; Zhang, L.; Chen, Z.; and Tang, J. 2021.
\newblock Learning to Iteratively Solve Routing Problems with Dual-Aspect Collaborative Transformer.
\newblock In Ranzato, M.; Beygelzimer, A.; Dauphin, Y.~N.; Liang, P.; and Vaughan, J.~W., eds., \emph{Advances in Neural Information Processing Systems 34: Annual Conference on Neural Information Processing Systems 2021, NeurIPS 2021, December 6-14, 2021, virtual}, 11096--11107.

\bibitem[{Mazyavkina et~al.(2021)Mazyavkina, Sviridov, Ivanov, and Burnaev}]{survey2}
Mazyavkina, N.; Sviridov, S.; Ivanov, S.; and Burnaev, E. 2021.
\newblock Reinforcement learning for combinatorial optimization: A survey.
\newblock \emph{Computers \& Operations Research}, 134: 105400.

\bibitem[{Min, Bai, and Gomes(2023)}]{min2023unsupervised_tsp}
Min, Y.; Bai, Y.; and Gomes, C.~P. 2023.
\newblock Unsupervised Learning for Solving the Travelling Salesman Problem.
\newblock \emph{ArXiv preprint}, abs/2303.10538.

\bibitem[{Miranda-Bront et~al.(2017)Miranda-Bront, Curcio, M{\'e}ndez-D{\'\i}az, Montero, Pousa, and Zabala}]{miranda2017dac_vrp4}
Miranda-Bront, J.~J.; Curcio, B.; M{\'e}ndez-D{\'\i}az, I.; Montero, A.; Pousa, F.; and Zabala, P. 2017.
\newblock A cluster-first route-second approach for the swap body vehicle routing problem.
\newblock \emph{Annals of Operations Research}, 253: 935--956.

\bibitem[{Nowak et~al.(2018)Nowak, Villar, Bandeira, and Bruna}]{nowak2018revised}
Nowak, A.; Villar, S.; Bandeira, A.~S.; and Bruna, J. 2018.
\newblock Revised note on learning quadratic assignment with graph neural networks.
\newblock In \emph{2018 IEEE Data Science Workshop (DSW)}, 1--5. IEEE.

\bibitem[{Pan et~al.(2023)Pan, Jin, Ding, Feng, Zhao, Song, and Bian}]{pan2023htsp}
Pan, X.; Jin, Y.; Ding, Y.; Feng, M.; Zhao, L.; Song, L.; and Bian, J. 2023.
\newblock H-TSP: Hierarchically Solving the Large-Scale Travelling Salesman Problem.
\newblock \emph{ArXiv preprint}, abs/2304.09395.

\bibitem[{Qiu, Sun, and Yang(2022)}]{dimes}
Qiu, R.; Sun, Z.; and Yang, Y. 2022.
\newblock DIMES: A Differentiable Meta Solver for Combinatorial Optimization Problems.
\newblock \emph{ArXiv preprint}, abs/2210.04123.

\bibitem[{Song et~al.(2023)Song, Mi, Li, Zhuang, and Cao}]{song2023drl_economic_lot_scheduling}
Song, W.; Mi, N.; Li, Q.; Zhuang, J.; and Cao, Z. 2023.
\newblock Stochastic Economic Lot Scheduling via Self-Attention Based Deep Reinforcement Learning.
\newblock \emph{IEEE Transactions on Automation Science and Engineering}.

\bibitem[{Sui et~al.(2021)Sui, Ding, Liu, Xu, and Bu}]{3opt}
Sui, J.; Ding, S.; Liu, R.; Xu, L.; and Bu, D. 2021.
\newblock Learning 3-opt heuristics for traveling salesman problem via deep reinforcement learning.
\newblock In \emph{Asian Conference on Machine Learning}, 1301--1316. PMLR.

\bibitem[{Sun and Yang(2023)}]{sun2023difusco}
Sun, Z.; and Yang, Y. 2023.
\newblock DIFUSCO: Graph-based Diffusion Solvers for Combinatorial Optimization.
\newblock \emph{ArXiv preprint}, abs/2302.08224.

\bibitem[{Taillard and Helsgaun(2019)}]{taillard2019popmusic4tsp}
Taillard, {\'E}.~D.; and Helsgaun, K. 2019.
\newblock POPMUSIC for the travelling salesman problem.
\newblock \emph{European Journal of Operational Research}, 272(2): 420--429.

\bibitem[{Vinyals, Fortunato, and Jaitly(2015)}]{PN}
Vinyals, O.; Fortunato, M.; and Jaitly, N. 2015.
\newblock Pointer Networks.
\newblock In Cortes, C.; Lawrence, N.~D.; Lee, D.~D.; Sugiyama, M.; and Garnett, R., eds., \emph{Advances in Neural Information Processing Systems 28: Annual Conference on Neural Information Processing Systems 2015, December 7-12, 2015, Montreal, Quebec, Canada}, 2692--2700.

\bibitem[{Wang et~al.(2021)Wang, Yang, Slumbers, Han, Guo, Zhang, and Wang}]{wang2021game}
Wang, C.; Yang, Y.; Slumbers, O.; Han, C.; Guo, T.; Zhang, H.; and Wang, J. 2021.
\newblock A Game-Theoretic Approach for Improving Generalization Ability of TSP Solvers.
\newblock \emph{ArXiv preprint}, abs/2110.15105.

\bibitem[{Wang et~al.(2023)Wang, Yu, McAleer, Yu, and Yang}]{wang2023asp}
Wang, C.; Yu, Z.; McAleer, S.; Yu, T.; and Yang, Y. 2023.
\newblock ASP: Learn a Universal Neural Solver!
\newblock \emph{ArXiv preprint}, abs/2303.00466.

\bibitem[{Williams(1992)}]{williams1992REINFORCE}
Williams, R.~J. 1992.
\newblock Simple statistical gradient-following algorithms for connectionist reinforcement learning.
\newblock \emph{Reinforcement learning}, 5--32.

\bibitem[{Wu et~al.(2021{\natexlab{a}})Wu, Song, Cao, and Zhang}]{wu2021learning_LNS_IP}
Wu, Y.; Song, W.; Cao, Z.; and Zhang, J. 2021{\natexlab{a}}.
\newblock Learning Large Neighborhood Search Policy for Integer Programming.
\newblock In Ranzato, M.; Beygelzimer, A.; Dauphin, Y.~N.; Liang, P.; and Vaughan, J.~W., eds., \emph{Advances in Neural Information Processing Systems 34: Annual Conference on Neural Information Processing Systems 2021, NeurIPS 2021, December 6-14, 2021, virtual}, 30075--30087.

\bibitem[{Wu et~al.(2021{\natexlab{b}})Wu, Song, Cao, Zhang, and Lim}]{wu2021learning}
Wu, Y.; Song, W.; Cao, Z.; Zhang, J.; and Lim, A. 2021{\natexlab{b}}.
\newblock Learning improvement heuristics for solving routing problems..
\newblock \emph{IEEE transactions on neural networks and learning systems}.

\bibitem[{Wu et~al.(2023)Wu, Zhou, Xia, Zhang, Cao, and Zhang}]{wu2023neural_airport}
Wu, Y.; Zhou, J.; Xia, Y.; Zhang, X.; Cao, Z.; and Zhang, J. 2023.
\newblock Neural Airport Ground Handling.
\newblock \emph{ArXiv preprint}, abs/2303.02442.

\bibitem[{Xiao et~al.(2019)Xiao, Zhang, Du, and Zhang}]{xiao2019dac_vrp3}
Xiao, J.; Zhang, T.; Du, J.; and Zhang, X. 2019.
\newblock An evolutionary multiobjective route grouping-based heuristic algorithm for large-scale capacitated vehicle routing problems.
\newblock \emph{IEEE transactions on cybernetics}, 51(8): 4173--4186.

\bibitem[{Xin et~al.(2021{\natexlab{a}})Xin, Song, Cao, and Zhang}]{MDAM}
Xin, L.; Song, W.; Cao, Z.; and Zhang, J. 2021{\natexlab{a}}.
\newblock Multi-Decoder Attention Model with Embedding Glimpse for Solving Vehicle Routing Problems.
\newblock In \emph{Thirty-Fifth {AAAI} Conference on Artificial Intelligence, {AAAI} 2021}, 12042--12049. {AAAI} Press.

\bibitem[{Xin et~al.(2021{\natexlab{b}})Xin, Song, Cao, and Zhang}]{neurolkh}
Xin, L.; Song, W.; Cao, Z.; and Zhang, J. 2021{\natexlab{b}}.
\newblock NeuroLKH: Combining Deep Learning Model with Lin-Kernighan-Helsgaun Heuristic for Solving the Traveling Salesman Problem.
\newblock In Ranzato, M.; Beygelzimer, A.; Dauphin, Y.~N.; Liang, P.; and Vaughan, J.~W., eds., \emph{Advances in Neural Information Processing Systems 34: Annual Conference on Neural Information Processing Systems 2021, NeurIPS 2021, December 6-14, 2021, virtual}, 7472--7483.

\bibitem[{Yang et~al.(2023)Yang, Zhao, Yuan, Yu, Li, and Gu}]{yang2023memory-efficient-transformer}
Yang, H.; Zhao, M.; Yuan, L.; Yu, Y.; Li, Z.; and Gu, M. 2023.
\newblock Memory-efficient Transformer-based network model for Traveling Salesman Problem.
\newblock \emph{Neural Networks}, 161: 589--597.

\bibitem[{Ye et~al.(2023)Ye, Wang, Cao, Liang, and Li}]{ye2023deepaco}
Ye, H.; Wang, J.; Cao, Z.; Liang, H.; and Li, Y. 2023.
\newblock DeepACO: Neural-enhanced Ant Systems for Combinatorial Optimization.
\newblock \emph{ArXiv preprint}, abs/2309.14032.

\bibitem[{Zhang et~al.(2022{\natexlab{a}})Zhang, Zhang, Cao, Song, Tan, Zhang, Wen, and Dauwels}]{zhang2022mtsptwr}
Zhang, R.; Zhang, C.; Cao, Z.; Song, W.; Tan, P.~S.; Zhang, J.; Wen, B.; and Dauwels, J. 2022{\natexlab{a}}.
\newblock Learning to solve multiple-TSP with time window and rejections via deep reinforcement learning.
\newblock \emph{IEEE Transactions on Intelligent Transportation Systems}.

\bibitem[{Zhang et~al.(2021)Zhang, Mei, Huang, Zheng, and Zhang}]{zhang2021dac_vrp1}
Zhang, Y.; Mei, Y.; Huang, S.; Zheng, X.; and Zhang, C. 2021.
\newblock A route clustering and search heuristic for large-scale multidepot-capacitated arc routing problem.
\newblock \emph{IEEE Transactions on Cybernetics}, 52(8): 8286--8299.

\bibitem[{Zhang et~al.(2022{\natexlab{b}})Zhang, Zhang, Wang, and Zhu}]{hac}
Zhang, Z.; Zhang, Z.; Wang, X.; and Zhu, W. 2022{\natexlab{b}}.
\newblock Learning to Solve Travelling Salesman Problem with Hardness-Adaptive Curriculum.
\newblock In \emph{Thirty-Sixth {AAAI} Conference on Artificial Intelligence, {AAAI} 2022}, 9136--9144. {AAAI} Press.

\bibitem[{Zhou et~al.(2023)Zhou, Wu, Song, Cao, and Zhang}]{zhou2023towards_Omni_generalizable}
Zhou, J.; Wu, Y.; Song, W.; Cao, Z.; and Zhang, J. 2023.
\newblock Towards Omni-generalizable Neural Methods for Vehicle Routing Problems.
\newblock In \emph{the 40th International Conference on Machine Learning (ICML 2023)}.

\bibitem[{Zong et~al.(2022{\natexlab{a}})Zong, Wang, Wang, Zheng, and Li}]{zong2022rbg}
Zong, Z.; Wang, H.; Wang, J.; Zheng, M.; and Li, Y. 2022{\natexlab{a}}.
\newblock RBG: Hierarchically Solving Large-Scale Routing Problems in Logistic Systems via Reinforcement Learning.
\newblock In \emph{Proceedings of the 28th ACM SIGKDD Conference on Knowledge Discovery and Data Mining}, 4648--4658.

\bibitem[{Zong et~al.(2022{\natexlab{b}})Zong, Zheng, Li, and Jin}]{zong2022mapdp}
Zong, Z.; Zheng, M.; Li, Y.; and Jin, D. 2022{\natexlab{b}}.
\newblock {MAPDP:} Cooperative Multi-Agent Reinforcement Learning to Solve Pickup and Delivery Problems.
\newblock In \emph{Thirty-Sixth {AAAI} Conference on Artificial Intelligence, {AAAI} 2022}, 9980--9988. {AAAI} Press.

\end{thebibliography}

\ifincludeappendix

\appendix

\section{Details of GLOP}\label{app: Details of GLOP}
\subsection{Pseudo code of solving (sub-)TSP}\label{app: pseudo code}
We present the pseudo code of solving (sub-)TSP in Algorithm \ref{alg}. Note that we can generate multiple initial solutions (i.e., $W>1$) and pick the best one after all revisions. In line 12, the strategy for shifting the decomposition point is to minimize the subtour overlap between revisions.
\begin{algorithm*}[]
    \caption{Solving (sub-)TSP}
	\begin{algorithmic}[1]
	    \STATE {\bfseries Input:} A TSP instance $tsp$ with $N$ nodes; trained Revisers and a list of their sizes $RS$ (e.g., $\{100,50,20 \}$); the number of initial tours $W$; revision iterations $I_n$, $\forall n \in RS$
	    \STATE {\bfseries Output:} The best solution $\bm{\pi}^*$
        \STATE Generate the initial tours: $\{
        \bm{\pi}^1,
        \ldots,
        \bm{\pi}^W
        \} 
        \gets RandomInsertion(tsp, W)$
        \FOR{ $n \in RS$}
        \STATE Initialize the decomposition point: $p \gets 0 $
        \STATE Calculate the number of subtours decomposed from a tour: $K_n=\lfloor \frac{N}{n} \rfloor$
        
        \FOR{$iter = 1 \to I_n$}
        
        \STATE $subtours \gets \{ 
        \{ \bm{\pi}^1_{p:p+n},\ldots, \bm{\pi}^1_{p+n(K_n-1):p+nK_n}   \},
        \ldots, 
        \{ \bm{\pi}^W_{p:p+n},\ldots, \bm{\pi}^W_{p+n(K_n-1):p+nK_n}   \}
        \}$
        \STATE Apply coordinate transformation and instance augmentation to $subtours$
        \STATE $ subtours \gets \mbox{Reviser-}n(subtours, tsp)$
        
        \STATE $\{
        \bm{\pi}^1,
        \ldots,
        \bm{\pi}^W
        \} \gets Composition(subtours)$ 
        
        \STATE Shift the decomposition point: $p \gets p + \max(1, \lfloor \frac{n}{I_n} \rfloor)$
        \ENDFOR
        \ENDFOR
        \STATE $\bm{\pi}^* = PickBest(tsp, \{
        \bm{\pi}^1,
        \ldots,
        \bm{\pi}^W
        \})$
	    
	\end{algorithmic}
	\label{alg}
\end{algorithm*}

\subsection{Coordinate transformation}\label{app: coordinate transformation}
Recall that we solve SHPPs with local policies parameterized by deep neural models, i.e., revisers, and the coordinates of a subtour are the inputs to a reviser. In the inference phase, we apply a coordinate transformation to improve the predictability and homogeneity of the model inputs. It facilitates training and benefits inference performance. Let $(x^\prime_i, y^\prime_i)$ denotes the coordinates of the $i$th node after transformation; $x_{\max}$, $x_{\min}$, $y_{\max}$, and $y_{\min}$ denote the bounds of an SHPP graph. Then the coordinate transformation is formulated as
\begin{equation}
\begin{aligned}
    x^\prime_i=\left\{\begin{array}{ll}
    {
    sc(x_i-x_{\min})
    } 
    \quad \mbox{if  } x_{\max}-x_{\min} > y_{\max}-y_{\min},
    \\
    {
    sc(y_i-y_{\min})
    } 
    \quad \mbox{otherwise},
    \end{array}  \right.
    \\
    y^\prime_i=\left\{\begin{array}{ll}
    {
    sc(y_i-y_{\min})
    } 
    & \mbox{if  } x_{\max}-x_{\min} > y_{\max}-y_{\min},
    \\
    {
    sc(x_i-x_{\min})
    } 
    & \mbox{otherwise},
    \end{array}  \right. 
\end{aligned}
\end{equation}
where the scale coefficient $sc= \frac{1}{\max(x_{\max}-x_{\min}, y_{\max}-y_{\min})}$. 

With the above Min-max Normalization and an optional graph rotation, we scale the x-axis coordinates to $[0,1]$ and set the y-axis lower bound to 0. Nevertheless, the revisers need to handle inputs of varied y-axis upper bound. It motivates us to develop the multi-distribution curriculum.

\subsection{Curriculum learning}
\paragraph{Stage 1: multi-distribution training}
The first curriculum stage trains all revisers using multi-distribution SHPPs. For each training instance, we first sample a y-axis upper bound: $y_{\max} \sim (0,1]$, then the instance: $(x,y) \sim [0,1]\times[0,y_{\max}]$, both uniformly.

\paragraph{Stage 2: collaborative training}
We first generate TSP instances with nodes sampled uniformly in $[0,1]^2$, then decompose their insertion-generated TSP tours into SHPPs. These SHPPs are used to fine-tune the first reviser. Each fine-tuned reviser is applied to infer its training instances, and the output subtours are decomposed into smaller-scale SHPPs to fine-tune the next reviser.

\subsection{Details of our global policy} \label{app: details of our global policy}
\paragraph{Constraints}
$\Theta$ in Eq. (\ref{eq: sample global solution}) requires specification for certain problems. For CVRP, we constrain $|\bm{\pi}|$ to the maximum number of vehicles by using the mask function suggested by \citet{hou2023_tam} and prohibit revisiting the same nodes or exceeding vehicle capacity. For PCTSP, we set $|\bm{\pi}|$ to 1 and prohibit revisiting the same nodes or violating the minimum prize constraint.

\paragraph{Inputs}\label{app: Inputs to global policy}
For the CVRP input graph, the node features are the normalized demand and polar coordinates w.r.t. the depot; the edge attributes are the relative Euclidean distance and polar angle. We sparsify the input graph by restricting each node to only connect with its $k$ nearest neighbors based on their polar angles. $k$ is set to 100 for CVRP1K and 200 for the rest. For the PCTSP input graph, the node features are the normalized prize and penalty; the edge attribute is the relative Euclidean distance. We sparsify the graph based on the Euclidean distance, setting $k$ to 50, 100, and 200 for PCTSP500, 1K, and 5K, respectively.

\subsection{Model architecture and training settings}

\paragraph{Local policy}
We deploy revisers of four scales: Reviser-100, 50, 20, and 10. The hyperparameter configurations of our revisers are the same as \cite{Attention2018}, except that we use 6 encoder layers instead of 3. We apply Adam optimizer \cite{Adam} ($\eta=10^{-4}$). The 1st-stage curriculum trains Reviser-10 for 100 epochs, while others for 200 epochs. An epoch contains 1.28M SHPP instances. The 2nd-stage curriculum trains all revisers for 300 epochs with a learning rate decay of 0.99. A training epoch goes through a training dataset of 1M SHPP instances.

\paragraph{Global policy}
We employ the same model architecture as \cite{dimes}, except that we set the dimension of node embedding to 48 on CVRP. We apply AdamW optimizer ($\eta=1\times 10^{-4} \mbox{ for PCTSP}; \eta=3\times 10^{-4} \mbox{ for CVRP}$) \cite{loshchilov2017AdamW} with Cosine Annealing scheduler \cite{loshchilov2016sgdr} and train each global policy with 5.12K instances.

\subsection{Inference settings}\label{app: inference settings}
\paragraph{TSP}
The used hyperparameters are gathered in Table \ref{tab: tsp inference hp}. For the cross-distribution evaluation on TSP100, we skip instance augmentation and directly treat TSP100 as SHPP100 when implementing Reviser-100.

\paragraph{CVRP}
Our global policies implement greedy decoding for CVRP. The hyperparameter settings for the local policies are displayed in Table \ref{tab: cvrp inference hp}.

\paragraph{PCTSP}
On PCTSP, our local policies apply both greedy and sampling decoding. For the latter, the number of sampled partitions is set to 10. As for the local policies, we deploy Reviser-100, 50, and 20, and conduct 10, 10, and 5 revisions, respectively.

\begin{table*}[!tb]
\centering
\begin{tabular}{l|ccccc|ccccc}
\toprule
 & \multicolumn{5}{c|}{GLOP for TSP} & \multicolumn{5}{c}{GLOP (more revisions) for TSP } \\ 
 & \multicolumn{1}{l}{$W$} & \multicolumn{1}{l}{$I_{10}$} & \multicolumn{1}{l}{$I_{20}$} & \multicolumn{1}{l}{$I_{50}$} & \multicolumn{1}{l|}{$I_{100}$} & \multicolumn{1}{l}{$W$} & \multicolumn{1}{l}{$I_{10}$} & \multicolumn{1}{l}{$I_{20}$} & \multicolumn{1}{l}{$I_{50}$} & \multicolumn{1}{l}{$I_{100}$} \\ \midrule
TSP100  & 35 & 5 & 10 & 10 & 20 & 140 & 5 & 10 & 10 & 20 \\
TSP500 & 1 & - & 5 & 25 & 20 & 10 & - & 5 & 25 & 20 \\
TSP1K & 1 & - & 5 & 25 & 20 & 10 & - & 5 & 25 & 20 \\
TSP10K & 1 & - & 5 & 20 & 10 & 1 & - & 5 & 25 & 50 \\
TSP100K & 1 & - & 5 & 5 & 5 & 1 & - & 5 & 25 & 50 \\ \bottomrule
\end{tabular}
\caption{Hyperparameter settings used for TSP inference.}
\label{tab: tsp inference hp}
\end{table*}

\begin{table}[t]
\centering
\begin{tabular}{l|ccc}
\toprule
 &$W$ & $I_{20}$ & $I_{50}$ \\ \midrule
CVRP1K & 1 & 5 & - \\
CVRP2K & 1 & 5 & 5 \\
CVRP5K & 1 & 5 & - \\
CVRP7K & 1 & 5 & - \\ \bottomrule
\end{tabular}
\caption{Hyperparameter settings used for CVRP inference.}
\label{tab: cvrp inference hp}
\end{table}

\section{Discussing two paradigms}\label{app: discussions of two paradigms}
From a unified perspective, both autoregressive (AR) and non-autoregressive (NAR) paradigms can be viewed as learning a probabilistic construction graph \cite{Generalization2022, goh2022combining_rl_optimal_transport}. 

The AR heuristics usually learn ``fine-grained'' construction graphs in the sense that, through forward passing of a neural network (NN) (e.g., attention mechanism using current context as query \cite{Attention2018}), per-step construction is heavily conditioned on the obtained partial solution. So it takes more meticulous actions by referring to the rich decoding context. However, it entails accurate context representation for effective actions and costly step-by-step neural decoding for sampling training trajectories, which hinder the scalability of AR heuristics \cite{Generalization2022}.

The NAR heuristics usually learn ``coarse-grained'' construction graphs only conditioned on the input problem instance. It enables construction graph (heatmap) generations in one shot and dramatically simplifies the context representation for solution decoding (e.g., decoding while simply masking the visited nodes). While being much more scalable, they underperform with vanilla sampling-based decoding due to the paucity of rich decoding context \cite{joshi2019GCN, dimes, sun2023difusco}. In this sense, those NAR heuristics need to incorporate additional solution refinements to trade off execution time for higher solution quality, making them less suitable for real-time applications.

\section{Additional related work} \label{app: additional related work}
The end-to-end NCO solvers can be categorized into two paradigms, i.e., autoregressive (AR) solution construction and non-autoregressive (NAR) heatmap generation coupled with subsequent decoding.

AR heuristics allow the neural models to output the assignments to decision variables sequentially, one step at a time \cite{PN, belloRL, Attention2018, ma2019HRL, POMO, MDAM, hottung2019nlns, bresson2021transformer_for_tsp, sgbs, symNCO, zong2022mapdp, grinsztajn2022poppy, zhang2022mtsptwr, jin2023pointerformer}. Learning global AR methods suffers from poor scaling-up generalization, as well as the prohibitive cost and ineffectiveness of training on large problems \cite{Generalization2022}. By contrast, NAR heuristics typically apply GNNs in one shot to output a heatmap predicting how promising each solution component is. Then, one can directly generate solutions with such a heatmap using greedy decoding, sampling, or beam search \cite{nowak2018revised, joshi2019GCN, dimes, sun2023difusco, ye2023deepaco}, which are all considered as vanilla sampling-based decoding strategies. Despite recent success in scaling NAR methods to large-scale problems \cite{fu2021generalize, dimes,sun2023difusco, min2023unsupervised_tsp}, they underperform with vanilla sampling-based decoding and have to be coupled with iterative solution refinement such as Monte Carlo Tree Search (MCTS) \cite{fu2021generalize}, falling short of real-time needs.

To the best of our knowledge, GLOP is the first neural solver to effectively integrate both AR and NAR components. Regarding TSP solution quality, GLOP is not in competition with the methods equipped with sophisticated and iterative improvement operators such as MCTS and LKH-3. Instead, it is a complement for scenarios requiring highly real-time solutions. Moreover, despite the recent preliminary endeavor to learn a cross-distribution TSP solver \cite{wang2021game, hac, dro, KD2022, wang2023asp, zhou2023towards_Omni_generalizable, jiang2023ensemble}, GLOP performs SOTA consistency across TSP scales and distributions.

Less related to GLOP, some works explore hybridizing heatmap with other algorithms instead of providing end-to-end solutions. They includes utilizing heatmaps to assist MCTS \cite{fu2021generalize}, Guided Local Search \cite{hudson2021gls}, Guided Tree Search \cite{2018gcn_guided_tree_search}, LKH-3 \cite{neurolkh}, and Dynamic Programming \cite{kool2022ddp}. Another line of methods learn repeated decision-making for algorithmic enhancement or solution refinement \cite{DACT, wu2021learning, 2opt, 3opt, zong2022rbg, wu2021learning_LNS_IP, L2I, ma2023neuopt}. While these methods can continuously improve the solution quality, they often lack scalability and fail to meet the requirements for real-time solutions. In addition, a surge of recent works broaden the applications of NCO techniques \cite{wu2023neural_airport, zong2022mapdp, chen2022_mixed_delivery_pickup, fan2022UAV, song2023drl_economic_lot_scheduling}.

\section{Problem definitions}\label{app: problem definitions}
\paragraph{TSP} Given a set of cities and the distances between each pair of cities, the objective of the Traveling Salesman Problem is to find the shortest possible tour that visits each city exactly once and returns to the original city. The tour must be a closed loop, and the total distance traveled along this tour, known as the tour length, should be minimized.
\paragraph{SHPP} The Shortest Hamiltonian Path Problem (SHPP) is also referred to as the open-loop Traveling Salesman Problem (TSP) \cite{pan2023htsp}. In the SHPP, the goal is to find the shortest Hamiltonian path in a given graph. Given fixed starting/terminating nodes (they can be different nodes), a Hamiltonian path visits all other nodes exactly once.
\paragraph{CVRP} In Capacitated Vehicle Routing Problem (CVRP), a set of customers, each with a specific demand for goods, must be serviced by a fleet of vehicles originating from a central depot. CVRP determines the optimal sequence in which all customers are visited and serviced, while ensuring that the sum of their demands along each route does not exceed the vehicle capacity.
\paragraph{PCTSP} The setup of Prize Collecting Traveling Salesman Problem (PCTSP) follows that in \cite{Attention2018}. In PCTSP, each node is associated with a penalty for not visiting it and a prize for visiting it. PCTSP aims to minimize the total tour length plus penalties for unvisited nodes, while ensuring collecting a minimum total prize.

\section{Extended evaluations}

\subsection{Further comparison with baselines using MCTS}
GLOP compares favorably with baselines that implement MCTS \cite{fu2021generalize, dimes} when instant solutions are desired. Specifically, GLOP can produce reasonable solutions before the baselines obtain heatmaps, while the latter can eventually outperform GLOP due to the continued improvements provided by MCTS. The trends are shown in Fig. \ref{fig: fc} using TSP10K for illustration.

\begin{figure}[!h]
\centering
\includegraphics[width=0.45\textwidth]{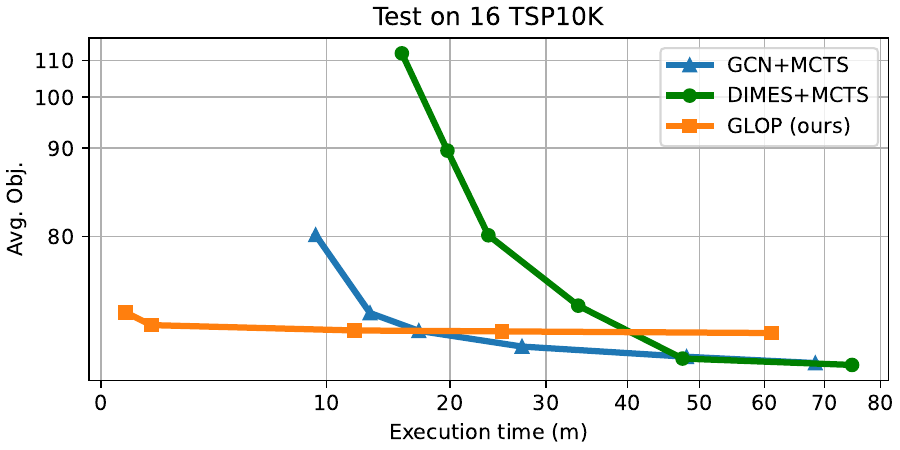}
\caption{Further comparison with two strong baselines that implement MCTS. The starting point of GLOP applies $W=1$ and no augmentation. The curves of GCN+MCTS and DIMES start when they finish heatmap generation and the first MCTS iteration.}
\label{fig: fc}
\end{figure}

\subsection{Ablation study}\label{sec: ablation}

\paragraph{Neural heuristics against LKH-3 in local policy}

We use neural heuristics as SHPP solvers for parallelizability and efficiency. Our local policies involve solving many SHPP instances, e.g., 32K SHPP20, 64K SHPP50, and 25.6K SHPP100 for 128 TSP1000, when using the hyperparameters (less revisions) in Table \ref{tab: tsp inference hp}. Neural heuristics can solve thousands of SHPPs simultaneously by leveraging the parallel computing power of GPUs, thereby improving efficiency. We evaluate neural heuristics against LKH-3 in our local policy to validate this motivation, with results on TSP and PCTSP displayed in Table \ref{tab: local policy validation}. Here, we replace the neural heuristics with LKH-3 (with the parameter `runs' set to 1 to save time). Implementing neural heuristics shows clear advantage in terms of solution efficiency.

\begin{table}[!htbp]
\centering
\resizebox{0.45\textwidth}{!}{
\begin{tabular}{c|cc|cc}
\toprule
 & \multicolumn{2}{c|}{w/ LKH-3} & \multicolumn{2}{c}{w/ neural heuristics (GLOP)} \\
 & Obj. & Time & Obj. & Time \\ \midrule
TSP500 & 16.86 & 34m & 17.07 & 19s \textbf{($\times$107)} \\
TSP1K & 23.74 & 68m & 24.01 & 34s \textbf{($\times$120)} \\
TSP10K & 74.64 & 50m & 75.62 & 32s \textbf{($\times$95)} \\
PCTSP500 & 14.5 & 17m & 14.6 & 26s \textbf{($\times$39)} \\
PCTSP1K & 19.8 & 25m & 20.0 & 47s \textbf{($\times$32)} \\
PCTSP5K & 45.7 & 1.7h & 46.0 & 3.7m \textbf{($\times$27)} \\ \bottomrule
\end{tabular}}
\caption{Neural heuristics against LKH-3 as SHPP solver.}
\label{tab: local policy validation}
\end{table}

\paragraph{Components in local policy}
We conduct ablation studies of the components in our inference pipeline for (sub-)TSP, including bidirectional decoding (BD), coordinate transformation (CT), and two-stage curriculum learning (CL1 and CL2). The results in Table \ref{tab: local ablation} validate the design of each component. Here, we use more revisions to realize the full potential of our local policies.

\paragraph{Global policy}
On CVRP, we demonstrate the advantage of our NAR global policy over AR dividing policy by comparing GLOP-G (LKH3) with TAM-LKH3 \cite{hou2023_tam} in Table \ref{tab: cvrp main results}. On PCTSP, Table \ref{tab: global ablation} further verifies the importance of our global policy by comparing it with random partition.

\begin{table}[]
\centering
\resizebox{0.35\textwidth}{!}{
\begin{tabular}{cccc|cc}
\toprule
\multicolumn{4}{c|}{Components} & \multirow{2}{*}{TSP500} & \multirow{2}{*}{TSP1K} \\ 
BD & CT & CL1 & CL2 &  &    \\ \midrule
$\times$ &  &  &   & 16.88 & 23.85 \\
 & $\times$ &  &   & 16.92 & 24.17 \\
 &  & $\times$ &  & 16.91 & 23.89 \\
 &  &  & $\times$  & 16.94 & 23.95 \\ \midrule
 &  &  &  & \textbf{16.80} & \textbf{23.73} \\ 
 \bottomrule
\end{tabular}}
\caption{Ablation studies of our local policy.}
\label{tab: local ablation}
\end{table}

\begin{table}[]
\centering
\begin{tabular}{l|cc}
\toprule
Problem & GLOP-RP & GLOP-G \\ \midrule
PCTSP500 & 19.0 & \textbf{14.6} \\
PCTSP1K & 26.1 & \textbf{20.0} \\
PCTSP5K & 53.0 & \textbf{46.0} \\ \bottomrule
\end{tabular}
\caption{Ablation studies of our global policy.}
\label{tab: global ablation}
\end{table}

\begin{table*}[]
\resizebox{\textwidth}{!}{
\begin{tabular}{l|ccccccc|cc}
\toprule
\multicolumn{1}{c|}{\multirow{2}{*}{\begin{tabular}[c]{@{}c@{}}Method\\ $\{ W=40\}$\end{tabular}}} & Uniform & \multicolumn{2}{c}{Expansion} & \multicolumn{2}{c}{Explosion} & \multicolumn{2}{c|}{Implosion} & \multicolumn{2}{c}{Avg.} \\
\multicolumn{1}{c|}{} & Gap(\%) & Gap(\%) & Det.(\%) & Gap(\%) & Det.(\%) & Gap(\%) & Det.(\%) & Gap(\%) & Det.(\%) \\ \midrule
AM+HAC $\{M=1280 \}$ & 2.484 & 3.997 & 61 & 3.084 & 24 & 2.595 & 4.5 & 3.040 & 30 \\ \midrule
GLOP (Uniform) & 0.215 & 0.235 & 9.3 & 0.119 & -44 & 0.197 & -8.4 & 0.192 & -14 \\
GLOP (C1) & 0.183 & 0.189 & 3.3 & 0.089 & -51 & 0.162 & -11 & 0.156 & -20 \\
GLOP (C2) & 0.180 & 0.260 & 44 & 0.105 & -42 & 0.159 & -12 & 0.176 & -3 \\
GLOP (C1 + C2) & 0.091 & 0.166 & 82 & 0.066 & -27 & 0.082 & -9.9 & 0.101 & 15 \\ 
\bottomrule
\end{tabular}}
\caption{The impact of different training schemes on the cross-distribution performance of GLOP. GLOP (Uniform), GLOP (C1), and GLOP (C2) are trained with uniformly sampled SHPPs, the 1st-stage curriculum, and the 2nd-stage curriculum, respectively.}
\label{tab: training scheme}
\end{table*}

\begin{table*}[!th]
\resizebox{\textwidth}{!}{
\begin{tabular}{l|ccccccc|cc}
\toprule
\multicolumn{1}{c|}{\multirow{2}{*}{\begin{tabular}[c]{@{}c@{}}Method\\ $\{ W=40\}$\end{tabular}}} & Uniform & \multicolumn{2}{c}{Expansion} & \multicolumn{2}{c}{Explosion} & \multicolumn{2}{c|}{Implosion} & \multicolumn{2}{c}{Avg.} \\
\multicolumn{1}{c|}{} & Gap(\%) & Gap(\%) & Det.(\%) & Gap(\%) & Det.(\%) & Gap(\%) & Det.(\%) & Gap(\%) & Det.(\%) \\ \midrule
AM+HAC $\{M=1280 \}$ & 2.484 & 3.997 & 61 & 3.084 & 24 & 2.595 & 4.5 & 3.040 & 30 \\ \midrule
GLOP ($RS=\{50\}$) & 0.380 & 0.740 & 94 & 0.285 & -25 & 0.362  & -4.8 & 0.442 & 22 \\
GLOP ($I_n=1$)     & 0.738 & 1.194 & 62 & 0.621 & -16 & 0.693 & -6.1 & 0.811 & 13 \\
GLOP             & 0.091 & 0.166 & 82  & 0.066 & -27 & 0.082 & -9.9 & 0.101 & 15 \\ 
\bottomrule
\end{tabular}
}
\caption{The impact of more revisers or more revisions on the cross-distribution performance of GLOP. GLOP ($RS=\{50\}$) implements Reviser-50 alone; GLOP ($I_n=1$) implements a single revision for all revisers.}
\label{tab:ablation-cd-n}
\end{table*}

\subsection{Discussing cross-distribution performance of local policies}\label{app: cd contributer}
\paragraph{Training schemes}
In Table \ref{tab: training scheme}, we investigate four training schemes: utilizing uniformly sampled SHPPs, only the 1st-stage curriculum, only the 2nd-stage curriculum, and the original scheme. Compared with the cross-distribution baseline \cite{hac}, GLOP shows lower Det. under all settings. It verifies that the holistic solution process of dividing and conquering, instead of a particular training scheme, is the main contributor to the cross-distribution performance. By comparing GLOP with different training schemes, we also conclude that the 1st-stage curriculum enhances cross-distribution performance while the 2nd-stage curriculum makes GLOP more specialized in solving uniform TSP. The results should not come as a surprise: GLOP learns to solve SHPP with varied y-axis upper bound in the 1st-stage curriculum, and we sample uniform TSPs to construct the training datasets in the 2nd-stage curriculum.

\paragraph{More revisers and more revisions}
Intuitively, it may make GLOP more distribution-invariant to implement various local policies with different specialties and perform more revisions by decomposing TSP tours differently. However, Table \ref{tab:ablation-cd-n} demonstrates that more revisers or more revisions can effectively reduce ``Gap'' but have less impact on ``Det.''. We further conclude that the holistic solution process of dividing and conquering, instead of the implementation of multiple revisers or revisions, is the main contributor to the cross-distribution performance.

\begin{figure*}[!thb]
\centering
\includegraphics[width=\textwidth]{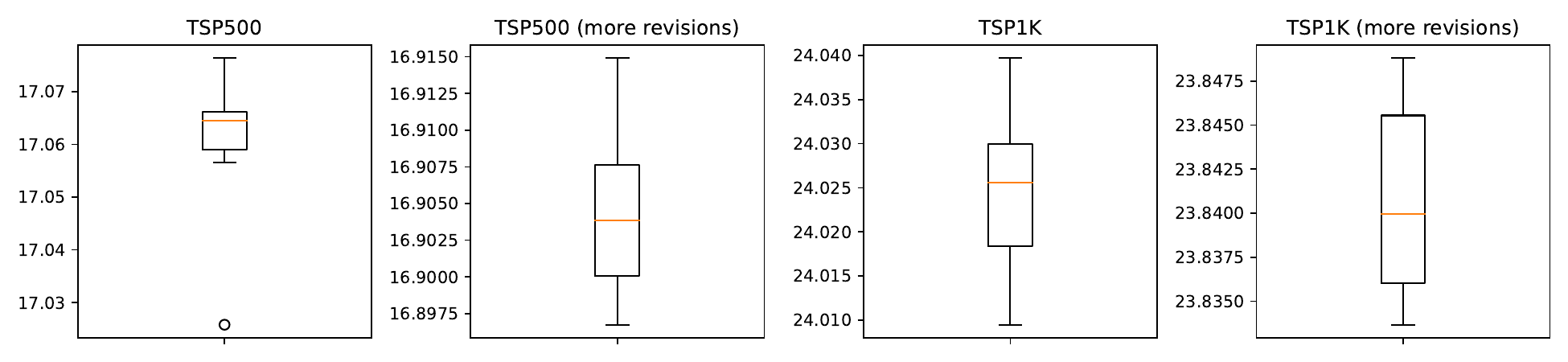}
\caption{Box plots of the objective values obtained by GLOP for 10
independent runs of 128 test instances (with random seeds 0-9).}
\label{fig: Stability analysis}
\end{figure*}

\subsection{Stability analysis of GLOP}\label{app: Stability analysis of GLOP}
Figure \ref{fig: Stability analysis} showcases the stability of GLOP on TSP500 and TSP1K, illustrating the box plots of the objective values for 10 independent runs. Each subplot depicts the minimum, lower quartile, mean, upper quartile, maximum, and possible outliers. The results suggest that implementing more revisions effectively guarantees better stability. The ranges of the box plots for GLOP (more revisions) are within 0.02, indicating desirable stability.

\subsection{Discussing time complexity}\label{app: discussing time complexity}

In GLOP, a revision decomposes a TSP into a batch of SHPPs and exploits NN to solve them simultaneously. Since the scale of SHPP is a constant for an arbitrary TSP, a revision enjoys linear time complexity. In practice, with full utilization of parallelism, the growth of execution time is slow. Table \ref{tab: time-complexity-tsp} presents the time needed for solving a single instance of different scales but under identical settings (50 and 20 revisions for Reviser-50, and 20, respectively). It shows a near-constant empirical complexity when the problem scale is less than 10K and the time of revisions dominates.

\begin{table*}[]
\centering
\begin{tabular}{c|cccccc}
\toprule
 Scale & TSP200 & TSP500 & TSP1K & TSP2K & TSP5K & TSP10K \\
 \midrule
Time (s) & 6.3 & 6.3 & 6.9 & 7.3 & 7.8 & 8.6 \\
\bottomrule
\end{tabular}
\caption{Time needed for solving a single instance of different scales.}
\label{tab: time-complexity-tsp}
\end{table*}

For CVRP/PCTSP, GLOP shows a (lower-than-)linear empirical complexity, as presented in Table \ref{tab: cvrp main results} and \ref{tab: exp-pctsp}, because the partition part has linear complexity due to graph sparsification and solving sub-TSPs has near-constant empirical complexity as discussed above.

\section{Implementation details of baselines}\label{app: Implementation details}
Our experiments involve the following neural baselines:
\begin{itemize}
    \item \textbf{TSP.} AM \cite{Attention2018}, LCP \cite{LCP}, GCN+MCTS \cite{fu2021generalize}, POMO-EAS \cite{eas2021}, DIMES \cite{dimes}, Tspformer \cite{yang2023memory-efficient-transformer}, H-TSP \cite{pan2023htsp}, Pointerformer \cite{jin2023pointerformer}, DACT \cite{DACT}, AMDKD+EAS \cite{KD2022}, 
    AM+HAC \cite{hac}, 
    and MatNet \cite{kwon2021matnet}.

    \item \textbf{CVRP.} AM \cite{Attention2018}, L2I \cite{L2I}, NLNS \cite{hottung2019nlns}, L2D \cite{delegate2021}, RBG \cite{zong2022rbg}, and TAM \cite{hou2023_tam}.

    \item \textbf{PCTSP.} AM \cite{Attention2018} and MDAM \cite{MDAM}.
\end{itemize}
Most baselines are reproduced by strictly following their open-sourced implementation. However, some others have to be tuned and adapted for proper and comparable evaluations.

\paragraph{Non-learning baselines} Most results of non-learning baselines are reproduced following \citet{Attention2018}. One exception is that we set the number of trails to 1 for LKH-3 on TSP100K to reduce runtime.

\paragraph{AM \cite{Attention2018}}
The checkpoints trained on TSP / PCTSP100 are generalized for comparison. We strictly follow the original implementation using sampling decoding.

\paragraph{POMO-EAS \cite{eas2021}}
We implement EAS-Tab using the POMO checkpoint trained on TSP100 and set $T=60$.

\paragraph{DACT \cite{DACT}}
For the TSPLIB benchmarks, we apply the model trained on TSP20 for instances with $<50$ nodes, the model trained on TSP50 for those with $50\leq n<100$ nodes, and the model trained on TSP100 for the rest, all with $T=1000$ and without augmentation.

\paragraph{GCN+MCTS \cite{fu2021generalize}}
We implement the CPU-version MCTS. The parameter $T$ that controls MCTS execution time is set smaller for a fair comparison. Note that $T$ can be directly derived from the total duration reported for MCTS. For example, on TSP500, it takes 33 seconds to complete MCTS for 128 instances on 12 CPU cores using multiprocessing (a single thread for each instance). Therefore, the time limit $T$ can be calculated as: 33s $\div$ 128 $\times$ 12. On TSPLIB benchmarks, we use the original TSP20 parameter settings for TSP20-49 instances, TSP50 for TSP50-99 instances, TSP100 for TSP100-149 instances, TSP200 for TSP150-349 instances, TSP500 for TSP350-749 instances, TSP1000 otherwise.

\paragraph{DIMES \cite{dimes}}
We follow the original implementations for TSP500, 1000, and 10000, except that we adjust the MCTS execution time $T$ to obtain comparable results. We generalize the model trained on TSP10K to TSP100K. For TSPLIB benchmarks, we additionally train a model on TSP50 and TSP200. We apply the TSP50 model and parameter settings for TSP20-199 instances, TSP200 for TSP200-349 instances, TSP500 for TSP350-749 instances, and TSP1000 otherwise.

\paragraph{AM+HAC \cite{hac}}
We fine-tune the AM trained on TSP100 with 100 epochs of Hardness-adaptive Curriculum (HAC) and set $M=1280$ for all evaluations.

\paragraph{AMDKD+EAS \cite{KD2022}}
We set the EAS iterations $T$ to 100.


\paragraph{L2D \cite{delegate2021} and RBG \cite{zong2022rbg}}
L2D and RBG adopt different CVRP settings: both set vehicle capacity to 50 for all scales, and RBG additionally fixes the depot to the center of the unit square. Their provided checkpoints are directly adapted for the evaluation in this work.

\paragraph{NLNS \cite{hottung2019nlns}}
We follow the original implementation of single-instance search and use the checkpoint trained on set XE\_17.

\paragraph{MDAM \cite{MDAM}}
The checkpoint trained on PCTSP100 is generalized for comparison.

\section{Discussing connections with non-learning OR methods}
NCO solvers show promise due to their parallelizability and data-driven nature in design automation, and they are gradually closing the gap with traditional OR methods. Following previous NCO works, GLOP serves as a versatile framework that effectively extends small neural solvers to much larger problems. As GLOP is founded on existing neural solvers, it stands to continually benefit from the rapid development of NCO models and increases in GPU power.

The idea of solving routing problems hierarchically has long been utilized in non-learning OR methods \cite{zhang2021dac_vrp1, alesiani2022dac_vrp2, xiao2019dac_vrp3, taillard2019popmusic4tsp}. A seminal work of this kind is POPMUSIC for TSP \cite{taillard2019popmusic4tsp}. Our (Sub-)TSP solver, as well as other hierarchical TSP solvers \cite{pan2023htsp, cheng23_select_and_optimize, LCP}, shares the ideology of POPMUSIC. Nevertheless, we discuss that NCO solvers and POPMUSIC have complementary natures. Employing the sophisticated initialization of POPMUSIC can improve the performance of NCO solvers. On the other hand, if replacing LKH with NCO solvers, POPMUSIC can gain from the high parallelizability of NCO at little cost of solution quality, given that current NCO solvers can solve small problems near-optimally.

\fi

\end{document}